\documentclass{sig-alternate-05-2015}
\usepackage{multirow}
\usepackage[utf8x]{inputenc}
\usepackage{natbib}
\usepackage{array}
\usepackage{graphicx}
\usepackage{float}
\usepackage{caption}
\usepackage{subcaption}

\begin{document}
\setcopyright{acmcopyright}
\setcopyright{acmlicensed}


\doi{10.475/123_4}

%

\conferenceinfo{KDD 2016}{August 13-17, San Francisco, CA, USA}
\acmPrice{\$15.00}

%

\title{Application of Deep Convolutional Neural Networks for Detecting Extreme Weather in Climate Datasets}

\numberofauthors{9} 
\author{
\alignauthor Yunjie Liu \\
       \affaddr{National Energy Research Scientific Computing Center}\\
       \affaddr{Lawrence Berkeley Lab}
       \affaddr{Berkeley, CA} \\
       \email{yunjieliu@lbl.gov}
\alignauthor Evan Racah\\
       \affaddr{National Energy Research Scientific Computing Center}\\
       \affaddr{Lawrence Berkeley Lab}
        \affaddr{Berkeley, CA} \\
       \email{eracah@lbl.gov}
\alignauthor Prabhat\\
       \affaddr{National Energy Research Scientific Computing Center}\\
       \affaddr{Lawrence Berkeley Lab}
       \affaddr{Berkeley, CA} \\
       \email{prabhat@lbl.gov}
\and   
\alignauthor Joaquin Correa\\
       \affaddr{National Energy Research Scientific Computing Center}\\
       \affaddr{Lawrence Berkeley Lab}
       \affaddr{Berkeley, CA} \\
       \email{joaquincorrea@lbl.gov}       
\alignauthor Amir Khosrowshahi\\
       \affaddr{Nervana Systems}\\
       \affaddr{San Diego, CA} \\
       \email{amir@nervanasys.com}
\alignauthor David Lavers\\
       \affaddr{Scripps Institution of Oceanography}\\
       \affaddr{San Diego, CA} \\
       \email{dlavers@ucsd.edu}
\and       
\alignauthor Kenneth Kunkel\\
       \affaddr{National Oceanic and Atmospheric Administration}\\
       \affaddr{Asheville, NC} \\
       \email{ken.kunkel@noaa.gov} 
\alignauthor Michael Wehner\\
       \affaddr{Lawrence Berkeley Lab}\\
       \affaddr{Berkeley, CA} \\
       \email{mfwehner@lbl.gov}
\alignauthor William Collins\\
       \affaddr{Lawrence Berkeley Lab}\\
       \affaddr{Berkeley, CA} \\
       \email{wdcollins@lbl.gov}
}
\date{Feb 12, 2016}

\maketitle

\begin{abstract}
Detecting extreme events in large datasets is a major challenge in climate science research. Current algorithms for extreme event detection are build upon human expertise in defining events based on subjective thresholds of relevant physical variables. Often, multiple competing methods produce vastly different results on the same dataset. Accurate characterization of extreme events in climate simulations and observational data archives is critical for understanding the trends and potential impacts of such events in a climate change content. This study presents the first application of Deep Learning techniques as alternative methodology for climate extreme events detection. Deep neural networks are able to learn high-level representations of a broad class of patterns from labeled data. In this work, we developed deep Convolutional Neural Network (CNN) classification system and demonstrated the usefulness of Deep Learning technique for tackling climate pattern detection problems. Coupled with Bayesian based hyper-parameter optimization scheme, our deep CNN system achieves 89\%-99\% of accuracy in detecting extreme events (Tropical Cyclones, Atmospheric Rivers and Weather Fronts). 
\end{abstract}




\section{Introduction}
Extreme climate events (such as hurricanes and heat waves) pose great potential risk on infrastructure and human health. Hurricane Joaquin, for example, hit Carolina in early October 2015, and dropped over 2 feet of precipitation in days, resulted in severe flooding and  economic loss. An important scientific goal in climate science research is to characterize extreme events in current day and future climate projections. However, understanding the developing mechanism and life cycle of these events as well as future trend requires accurately identifying such pattern in space and time. Satellites acquire 10s of TBs of global data every year to provide us with insights into the evolution of the climate system. In addition, high resolution climate models produces 100s of TBs of data from multi-decadal run to enable us to explore future climate scenarios under global warming. Detecting extreme climate events in terabytes of data presents an unprecedented challenge for climate science.
 
Existing extreme climate events (e.g. hurricane) detection methods all build upon human expertise in defining relevant events based on evaluating of relevant spatial and temporal variables on hard and subjective thresholds. For instance, tropical cyclones are strong rotating weather systems that are characterized by low pressure, warm temperature core structures with high wind. However, there is no universally accepted sets of criteria for what defines a tropical cyclone \cite{Nolan2012tropical}. The "Low" Pressure and "Warm" Temperature are interpreted differently among climate scientists, therefore different thresholds are used to characterize them.  Researchers \cite{vitart1997simulation,vitart1999impact,walsh1997tropical,walsh2007objectively,Prabhat2012teca,Prabhat2015teca} have developed various algorithms to detect tropical cyclones in large climate dataset based on subjective thresholding of several relevant variables (e.g. sea level pressure, temperature, wind etc.). One of the general and promising extreme climate event detecting software, Toolkit for Extreme Climate Analysis (TECA) \cite{Prabhat2012teca,Prabhat2015teca}, is able to detect tropical cyclones, extra-tropical cyclones and atmospheric rivers. TECA utilizes the MapReduce paradigm to find pattern in Terabytes of climate data with in hours. However, many climate extreme events do not have a clear empirical definition that is accepted universally by climate scientists (e.g. extra-tropical cyclone and mesoscale convective system), which precludes the development and application of algorithms for detection and tracking. This study attempts to search for an alternative methodology for extreme events detection by designing a neural network based system that is capable of learning a broad class of patterns from complex multi-variable climate data and avoiding subjective threshold.

Recent advances in deep learning have demonstrated exciting and promising results on pattern recognition tasks, such as ImageNet Large Scale Visual Recognition Challenge \cite{krizhevsky2012imagenet,Simonyan14c,szegedy2015going} and speech recognition \cite{hinton2012deep,dahl2012context,graves2013speech,sutskever2014sequence}.  Many of the state-of-art deep learning architectures for visual pattern recognition are based on the hierarchical feature learning convolutional neural network (CNN). Modern CNN systems tend to be deep and large with many hidden layers and millions of neurons, making them flexible in learning a broad class of patterns simultaneously from data. AlexNet (7 layers with 5 convolutonal layer and 2 fully connected layer) developed by \cite{krizhevsky2012imagenet} provides the first end to end trainable deep learning system on objective classification, which achieved 15.3\% top-5 classification error rate on ILSVRC-2012 data set. On the contrary, previous best performed non-neural network based systems achieved only 25.7\% top-5 classification error on the same data set. Shortly after that, Simonyan and Zisserman \cite{Simonyan14c} further developed AlexNet and introduced an even deeper CNN (19 layers with 16 convolutional layer and 3 fully connected layer) with smaller kernel (filter) and achieved an impressively 6.8\% top-5 classification error rate on ILSVRC-2014 data set. Szegedy et al.\cite{szegedy2015going} introduced the “inception” neural network concept (network includes sub-network) and developed an even deeper CNN (22 layers) that achieved comparable classification results on ImageNet benchmark. Build on deep CNN, Sermanet et al. \cite{sermanet14over} introduced an integrated system of classification and detection, in which features learned by convolutional layers are shared among classification and localization tasks and both tasks are performed simultaneously in a single network. Girshick et al. \cite{girshick2014rich} took a completely different approach by combining a region proposal framework \cite{uijlings2013selective} with deep CNN and designed the state of art R-CNN object detection system. 
 
In this paper, we formulate the problem of detecting extreme climate events as classic visual pattern recognition problem. We then build end to end trainable deep CNN systems, following the architecture introduced by \cite{krizhevsky2012imagenet}. The model was trained to classify tropical cyclone, weather front and atmospheric river. Unlike the ImageNet challenge, where the training data are labeled natural images, our training data consist of several continuous spatial variables(e.g. pressure, temperature, precipitation) and are stacked together into image patches.

\bigskip
\section{Related Work}
Climate data analysis requires an array of advanced methodology. Neural network based machine learning approach, as a generative analysis technique, has received much attention and been applied to tackle several climate problems in recent year. Chattopadhyay et al. \cite{chattopadhyay2013description} developed a nonlinear clustering method based on Self Organizational Map (SOM) to study the structure evolution of Madden–Julian oscillation (MJO). Their method does not require selecting leading modes or intraseasonal bandpass filtering in time and space like other methods do. The results show SOM based method is not only able to capture the gross feature in MJO structure and development but also reveals insights that other methods are not able to discover such as the dipole and tripole  structure of outgoing long wave radiation and diabatic heating in MJO. Gorricha and Costa \cite{gorricha2013framework} used a three dimensional Self Organizational Map on categorizing and visualizing extreme precipitation patterns over an island in Spain. They found spatial precipitation patterns that traditional precipitation index approach is not able to discover, and concluded that three dimensional Self Organizational Map is very useful tool on exploratory spatial pattern analysis. More recently, Shi et al. \cite{shi2015convolutional} implemented a newly developed convolutional long short term memory (LSTM) deep neural network for precipitation nowcasting. Trained on two dimensional radar map time series, their system is able to outperform the current state-of-art precipitation nowcasting system on various evaluation metrics. Iglesias et al. \cite{iglesiasexamination} developed a multitask deep fully connected neural network on prediction heat waves trained on historical time series data. They demonstrate that neural network approach is significantly better than linear and logistic regression. And potentially can improve the performance of forecasting extreme heat waves. These studies show that neural network is a generative method and can be applied on various climate problems. In this study, we explore deep Convolutional Neural Network on solving climate pattern detection problem.

\section{Methods}
\subsection{Convolutional Neural Network}
A Deep CNN is typically comprised of several convolutional layers followed by a small amount of fully connected layers. In between two successive convolutional layers, subsampling operation (e.g. max pooling, mean pooling) is performed typically. Researchers have argued about the necessity of pooling layers, and argue that they can be simply replaced by convolutional layer with increased strides, thus simplify the network structure \cite{SprDosBroRied15}. In either case, the inputs of a CNN is \textit{(m,n,p)} images, where \textit{m} and \textit{n} is the width and height of an image in pixel, \textit{p} is the number of color channel of each pixel. The output of a CNN is a vector of  q probability units (class scores), corresponding to the number of categories to be classified (e.g. for binary classifier \textit{q}=2).

The convolutional layers perform convolution operation between kernels and the input images (or feature maps from previous layer). Typically, a convolutional layer contains k filters (kernels) with the size \textit{(i,j,p)}. Where \textit{i, j} is the width and height of the filter. The filters  are usually smaller than the width \textit{m} and height \textit{n} of input image. \textit{p} always equal to the number of color channel of input image (e.g. a color image has three channels: red, green, and blue). Each of the filters is independently convolved with the input images (or feature maps from previous layer) followed by non-linear transformation and generates \textit{k} feature maps, which serve as inputs for the next layer. In the process of convolution, a dot product is computed between the entry of filter and the local region that it is connected to in the input image (or feature map from previous layer). The parameters of convolutional layer are these learnable filters. The convolutional layer is the feature extractor, because the kernels slide across all the inputs and will produce larger outputs for certain sub-regions than for others. This allows features to be extracted from inputs and preserved in the feature maps, which are passed on to next layer, regardless of where the feature is located in the input. The pooling layer subsamples the feature maps generated from convolutional layer over a \textit{(s,t)} contiguous region, where \textit{s, t} is the width and height of the subsampling window. This results in the resolution of the feature maps becoming coarser with the depth of CNN. All feature maps are high-level representations of the input data in CNN. The fully connected layer has connections to all hidden units in previous layer. If it is the last layer within CNN architecture, the fully connected layer also does the high level reasoning based on the feature vectors from previous layer and produce final class scores for image objects.

Most of current deep neural network uses back propagation as learning rule \cite{ruhmelhart1986learning}. The back propagation algorithm searches for minimum of loss function in weight space through gradient descent method.It partitions the final total loss to each of the single neuron in the network and repeatedly adjusts the weights of neurons whose loss is high, and back propagate the error through the entire network from output to its inputs.
 
 \subsection{Hyper-parameter Optimization}
Training deep neural network is known to be hard \cite{larochelle2009exploring,glorot2010understanding}. Effectively and efficiently train deep neural network not only requires large amount of training data, but also requires carefully tuning model hyper-parameters (e.g. learning parameters, regularization parameters) \cite{snoek2012practical}. The parameter tuning process, however, can be tedious and non-intuitive. Hyper-parameter optimization can be reduced to find a set of parameters for a network that produces the best possible validation performance. As such, this process can be thought of as a typical optimization problem of finding a set, $x$, of parameter values from a bounded set $X$ that minimize an objective function $f(x)$, where $x$ is a particular setting of the hyper-parameters and $f(x)$ is the loss for a deep neural network with a particular set of training and testing data as function of the hyper-parameter inputs. Training a deep neural network is not only a costly (with respect to time) procedure, but a rather opaque process with respect to how the network performance varies with respect to its hyper-parameter inputs. Because training and validating a deep neural network is very complicated and expensive, Bayesian Optimization (which assumes $f(x)$ is not known, is non-convex and is expensive to evaluate) is a well-suited algorithm for hyper-parameter optimization for our task at hand. Bayesian Optimization attempts to optimize $f(x)$ by constructing two things: a probabilistic model of $f(x)$ and an acquistion function that picks which point $x$ in $X$ to evaluate next. The probabilistic model is updated with Baye’s rule with a Gaussian prior. The acquisition function suggests hyper-parameter settings or points to evaluate by trying to balance evaluating parameter settings in regions, where $f(x)$ is low and points in regions where the uncertainty in the probabilistic model is high. As a result the optimization procedure attempts to evaluate as few points as possible \cite{brochu2010tutorial} \cite{snoek2012practical}. 
In order to implement Bayesian Optimization, we use a tool called Spearmint. Spearmint works by launching a Spearmint master process, which creates a database for collecting all model evaluation results. The master process then spawns many processes, which execute training and evaluation with respect to a set of hyper-parameters proposed by the acquisition function and then report their results to the database. From there, the master process uses the results in the database to propose further parameter settings and launch additional processes.

\subsection{CNN Configuration}
Following AlexNet \cite{krizhevsky2012imagenet}, we developed a deep CNN which has totally 4 learnable layers, including 2 convolutional layers and 2 fully connected layers. Each convolutional layer is followed by a max pooling layer. The model is constructed based on the open source python deep learning library NOEN. The configuration of our best performed architectures are shown in Table 1.

The networks are shallower and smaller comparing to the state-of-art architecture developed by \cite{Simonyan14c,szegedy2015going}.The major limitations for exploring deeper and larger CNNs is the limited amount of labeled training data that we can obtain. However, a small network has the advantage of avoiding over-fitting, especially when the amount of training data is small. We also chose comparatively large kernels (filters) in the convolutional layer based on input data size, even though \cite{Simonyan14c} suggests that deep architecture with small kernel (filter) is essential for state of art performance. This is because climate patterns are comparatively simpler and larger in size as compared to objects in ImageNet dataset.

One key feature of deep learning architectures is that it is able to learn complex non-linear functions. The convolutional layers and first fully connected layer in our deep CNNs all have Rectified Linear Unit (ReLU) activation functions \cite{nair2010rectified} as characteristic. ReLU is chosen due to its faster learning/training character \cite{krizhevsky2012imagenet} as compared to other activation functions like tanh.
              	\begin{equation}
                f(x)=max(0, x)
                \end{equation}
Final fully connected layer has Logistic activation function as non-linearity, which also serves as classifier and outputs a probability distribution over class labels.
              	\begin{equation}
                f(x)=\frac{1}{1+e^{-x}}
                \end{equation}
 
\begin{table*}
\centering
\captionsetup{width=0.7\textwidth}
\caption{Deep CNN architecture and layer parameters. The convolutional layer parameters are denoted as <filter size>-<number of feature maps> (e.g. 5x5-8). The pooling layer parameters are denoted as <pooling window> (e.g. 2x2).  The fully connected layer parameter are denoted as <number of units> (e.g. 2).}
\begin{tabular}{|c|p{1.6cm}|p{1.6cm}|p{1.6cm}|p{1.6cm}|p{1.3cm}|l|} \hline
 &Conv1&Pooling&Conv2&Pooling&Fully&Fully\\ \hline
 Tropical Cyclone&5x5-8&2x2&5x5-16&2x2&50&2\\ \hline
Weather Fronts&5x5-8&2x2&5x5-16&2x2&50&2\\ \hline
Atmospheric River&12x12-8&3x3&12x12-16&2x2&200&2\\ \hline
\end{tabular}
\end{table*}
\medskip

\begin{table*}
\centering
\captionsetup{width=0.7\textwidth}
\caption{Data Sources}
\begin{tabular}{|c|p{2.3cm}|p{3.2cm}|p{3.0cm}|} \hline
Climate Dataset&Time Frame&Temporal Resolution&Spatial Resolution (lat x lon degree) \\ \hline
CAM5.1 historical run&1979-2005&3 hourly&0.23x0.31\\ \hline
ERA-Interim reanalysis&1979-2011&3 hourly&0.25x0.25\\ \hline
20 century reanalysis&1908-1948&Daily&1x1\\ \hline
NCEP-NCAR reanalysis&1949-2009&Daily&1x1\\ \hline
\end{tabular}
\end{table*}

\begin{table*}[!ht]
\centering
\captionsetup{width=0.7\textwidth}
\caption{Size of image patch, diagnostic variables and number of labeled dataset used for extreme event considered in the study}
\begin{tabular}{|c|p{3.0 cm}|p{3.0cm}|p{3.3cm}|} \hline
Events&Image Dimension&Variables&Total Examples\\ \hline
Tropical Cyclone&32x32&PSL,VBOT,UBOT, T200,T500,TMQ, V850,U850&10,000 +ve 10,000 -ve\\ \hline
Atmospheric River&148 x 224&TMQ,Land Sea Mask&6,500 +ve 6,800 -ve \\ \hline
Weather Front &27 x 60&2m Temp, Precip, SLP&5,600 +ve 6,500 -ve\\ \hline
\end{tabular}
\end{table*}

\subsection{Computational Platform}
We performed our data processing, model training and testing on Edison, a Cray XC30 and Cori, a Cray XC40 supercomputing systems at the National Energy Research Scientific Computing Center (NERSC). Each of Edison computing node has 24 2.4 GHz Intel Xeon processors. Each of Cori computing node has 32 2.3 GHz Intel Haswell processors. In our work, we mainly used single node CPU backend of NEON. The hyper-parameter optimization was performed on a single node on Cori with tasks fully parallel on 32 cores.

\bigskip
\section{Data}
In this study, we use both climate simulations and reanalysis products. The reanalysis products are produced by assimilating observations into a climate model. A summary of the data source and its temporal and spatial resolution is listed in Table 2. Ground truth labeling of various events is obtained via multivariate threshold based criteria implemented in TECA \cite{Prabhat2012teca,Prabhat2015teca}, and manual labeling by experts \cite{kunkel2012meteorological,lavers2012detection}. Training data comprise of image patterns, where several relevant spatial variables are stacked together over a prescribed region that bounds a type of event. The dimension of the bounding box is based domain knowledge of events spatial extent in real word. For instance, tropical cyclone radius are typically with in range of 100 kilometers to 500 kilometers, thus bounding box size of 500 kilometers by 500 kilometers is likely to capture most of tropical cyclones. The chosen physical variables are also based on domain expertise. The prescribed bounding box is placed over the event. Relevant variables are extracted within the bounding box and stacked together. To facilitate model training, bounding box location is adjusted slightly such that all of events are located approximately at the center. Image patches are cropped and centered correspondingly. Because of the spatial dimension of climate events vary quite a lot and the spatial resolution of source data is non-uniform, final training images prepared differ in their size among the three types of event. A summary of the attributes of training images is listed in Table 3.

\section{Results and Discussion}
Table 4 summarizes the performance of our deep CNN architecture on classifying tropical cyclones, atmospheric rivers and weather fronts. We obtained fairly high accuracy (89\%-99\%) on extreme event classification. In addition, the systems do not suffer from over-fitting. We believe this is mostly because of the shallow and small size of the architecture (4 learnable layers) and the weight decay regularization. Deeper and larger architecture would be inappropriate for this study due to the limited amount of training data. Fairly good train and test classification results also suggest that the deep CNNs we developed are able to efficiently learn representations of climate pattern from labeled data and make predictions based on feature learned. Traditional threshold based detection method requires human expert carefully examine the extreme event and its environment, thus come up with thresholds for defining the events. In contrast, as shown in this study, deep CNNs are able to learn climate pattern just from the labeled data, thus avoiding subjective thresholds. 

\begin{table}[h]
\centering
\caption{Overall Classification Accuracy}
\begin{tabular}{|c|p{1.2cm}|p{1.2cm}|p{1.4cm}|} 
\hline
Event Type&Train &Test &Train time\\ 
\hline
Tropical Cyclone&99\% &99\% & $\approx$ 30 min\\ 
\hline
Atmospheric River&90.5\% &90\% & 6-7 hour \\ 
\hline
Weather Front&88.7\%&89.4\% & $\approx$ 30 min\\ 
\hline
\end{tabular}
\end{table}

\subsection{Classification Results for Tropical Cyclones}
Tropical cyclones are rapid rotating weather systems that are characterized by low pressure center with strong wind circulating the center and warm temperature core in upper troposphere. Figure 1 shows examples of tropical cyclones simulated in climate models, that are correctly classified by deep CNN (warm core structure is not shown in this figure). Tropical cyclone features are rather well defined, as can be seen from the distinct low pressure center and spiral flow of wind vectors around the center. These clear and distinct characteristics make tropical cyclone pattern relatively easy to learn and represent within CNN. Our deep CNNs achieved nearly perfect (99\%) classification accuracy. 

Figure 2 shows examples of tropical cyclones that are mis-classified. After carefully examining these events, we believe they are weak systems (e.g. tropical depression), whose low pressure center and spiral structure of wind have not fully developed. The pressure distribution shows a large low pressure area without a clear minimum. Therefore, our deep CNN does not label them as strong tropical cyclones.

\begin{table}[!h]
\centering
\caption{Confusion matrix for tropical cyclone classification}
\begin{tabular}{r|c|c|} 
\multicolumn{1}{r}{} & \multicolumn{1}{c}{Label TC} & \multicolumn{1}{c}{Label Non\_TC} \\
\cline{2-3}
Predict TC & 0.989 & 0.003 \\ 
\cline{2-3}
Predict Non\_TC&0.011& 0.997 \\ 
\cline{2-3}
\end{tabular}
\end{table}

\begin{figure}[!h]
\centering
\begin{subfigure}[b]{0.5\textwidth}
  \includegraphics[width=0.31\textwidth]{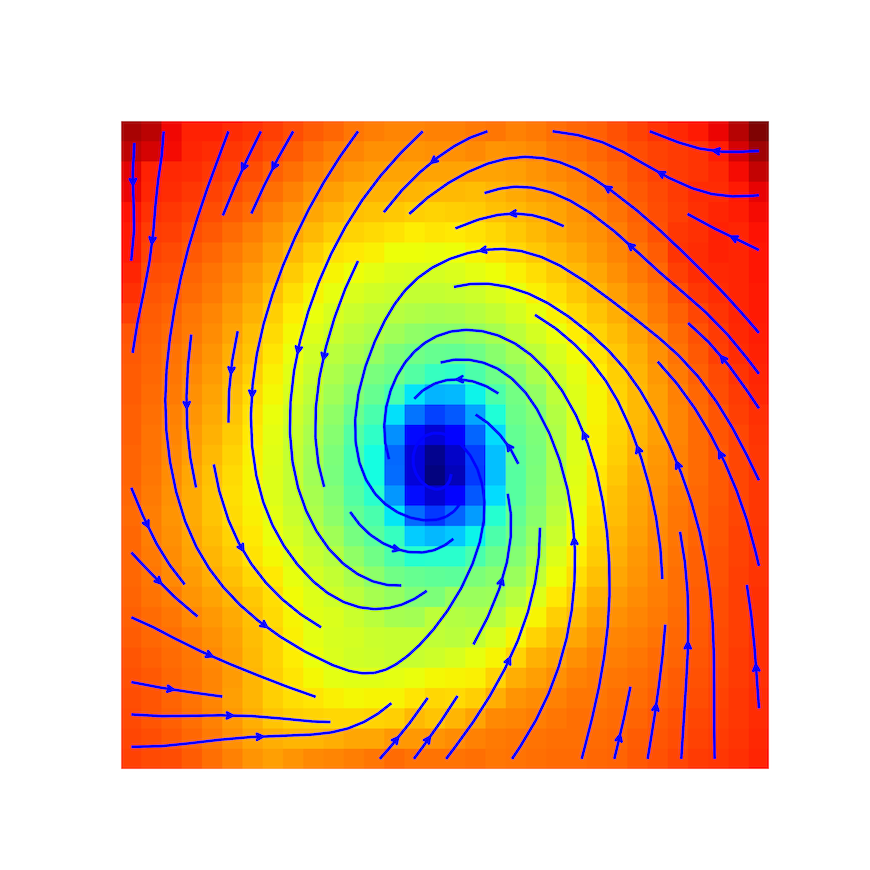}
  \includegraphics[width=0.31\textwidth]{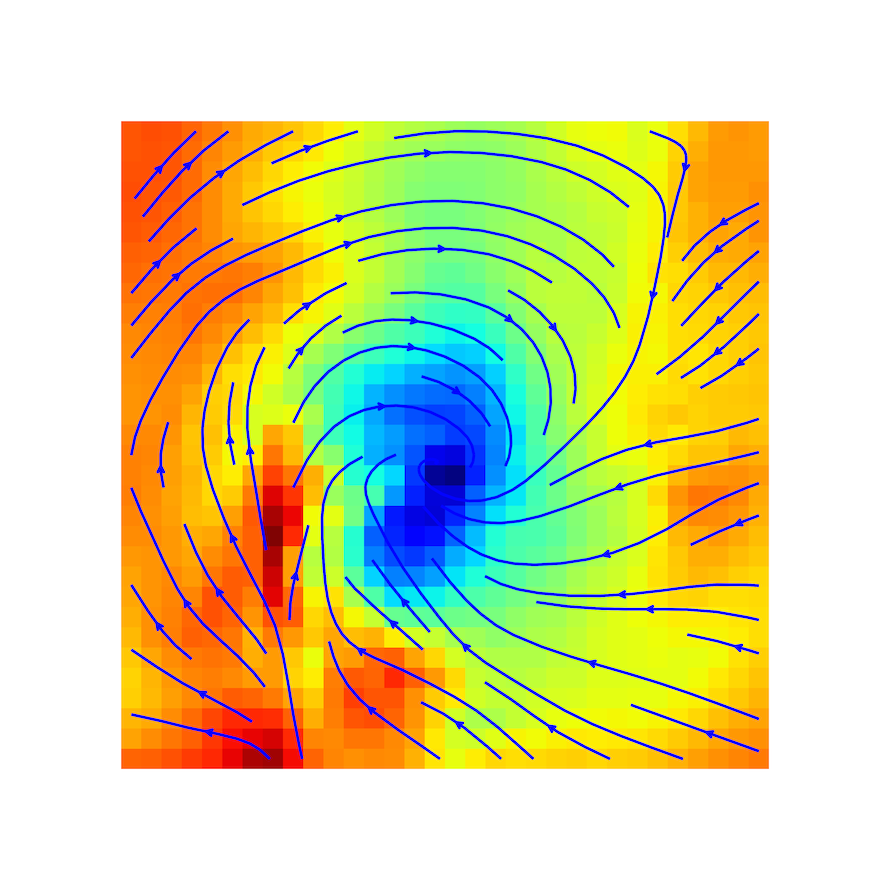}
  \includegraphics[width=0.31\textwidth]{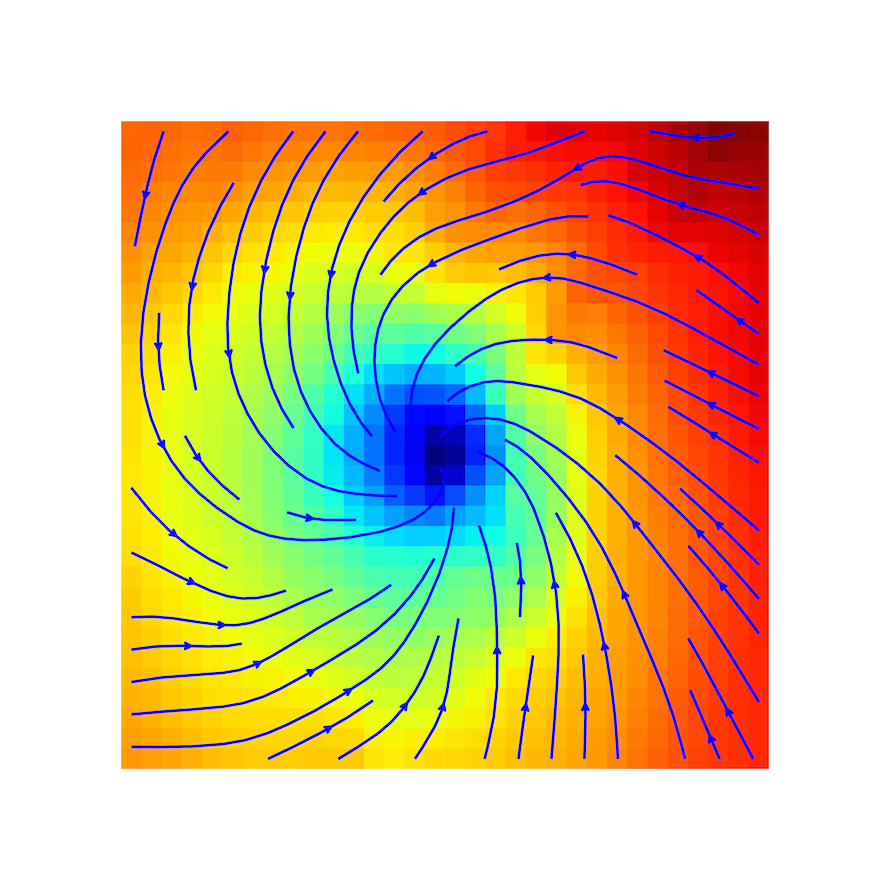}
\end{subfigure}
\begin{subfigure}[b]{0.5\textwidth}
  \includegraphics[width=0.31\textwidth]{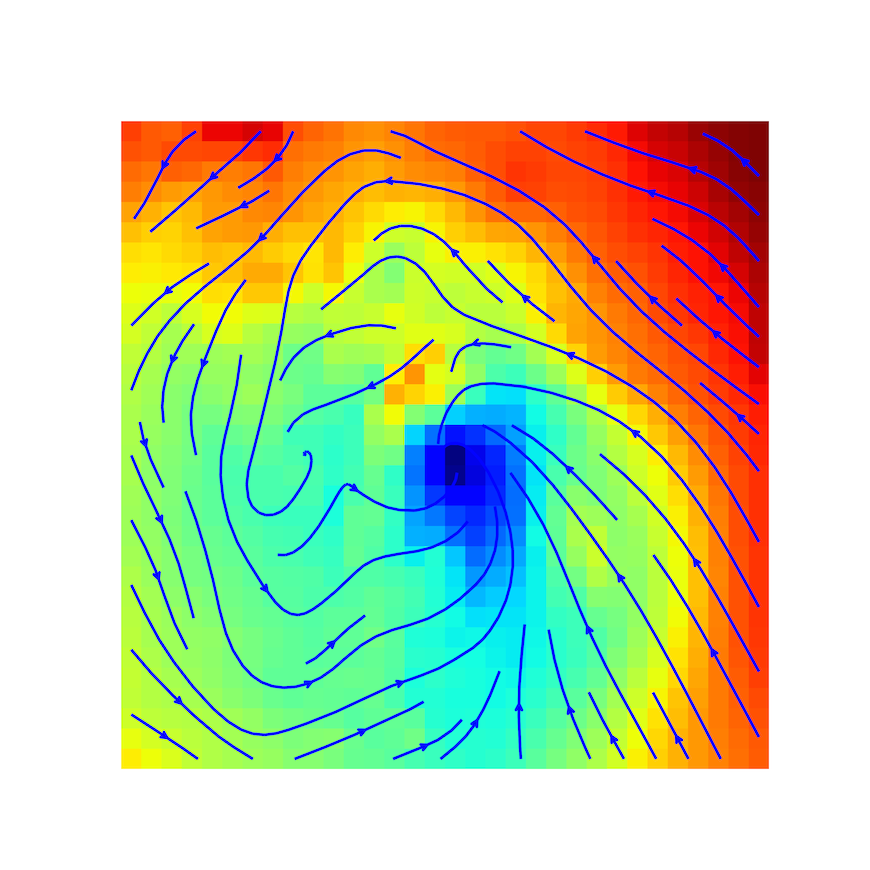}
  \includegraphics[width=0.31\textwidth]{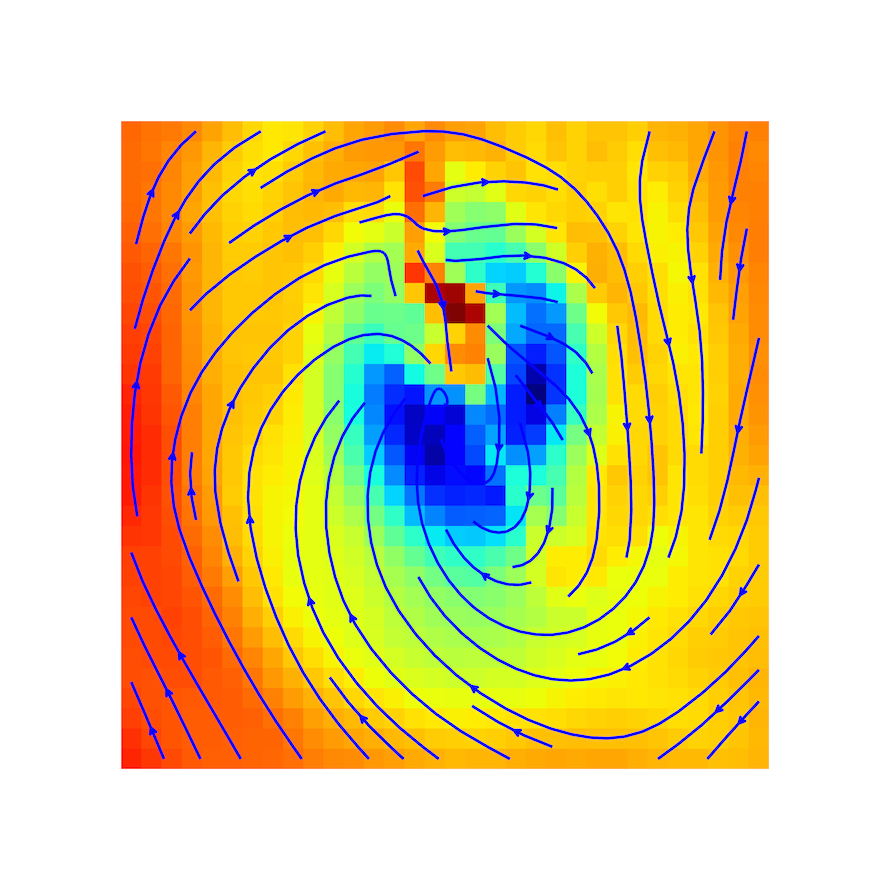}
  \includegraphics[width=0.31\textwidth]{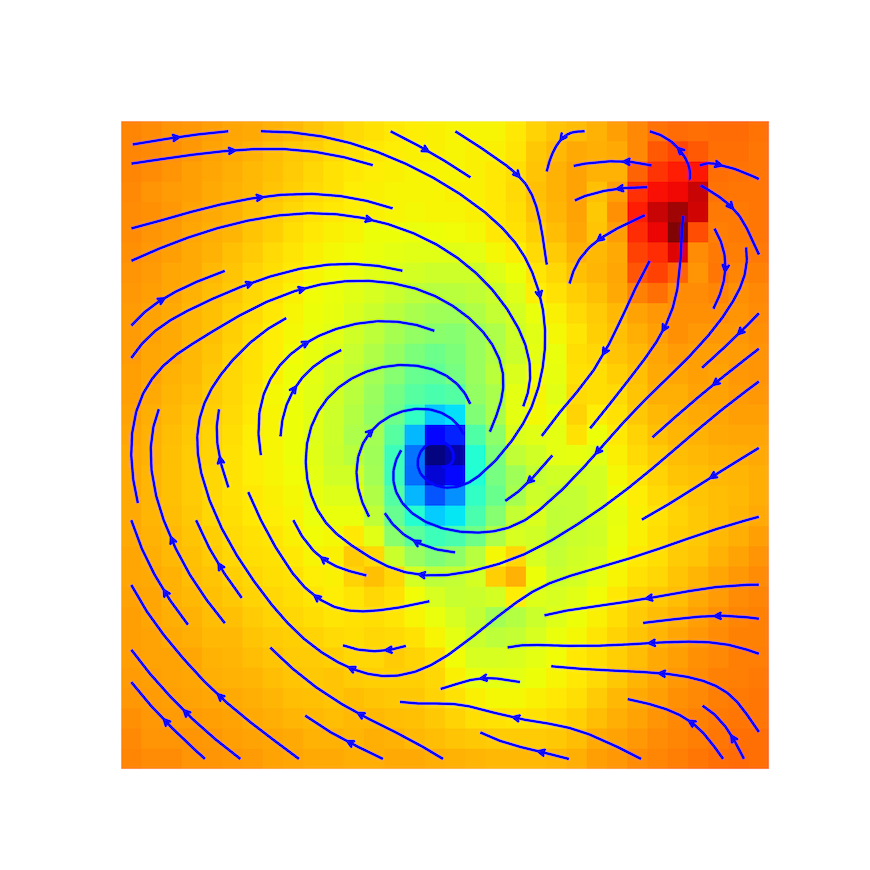}
\end{subfigure}
\caption{Sample images of tropical cyclones correctly classified (true positive) by our deep CNN model. Figure shows sea level pressure (color map) and near surface wind distribution (vector solid line).}
\end{figure}

\begin{figure}[!h]
\centering
\begin{subfigure}[b]{0.5\textwidth}
 \includegraphics[width=0.31\textwidth]{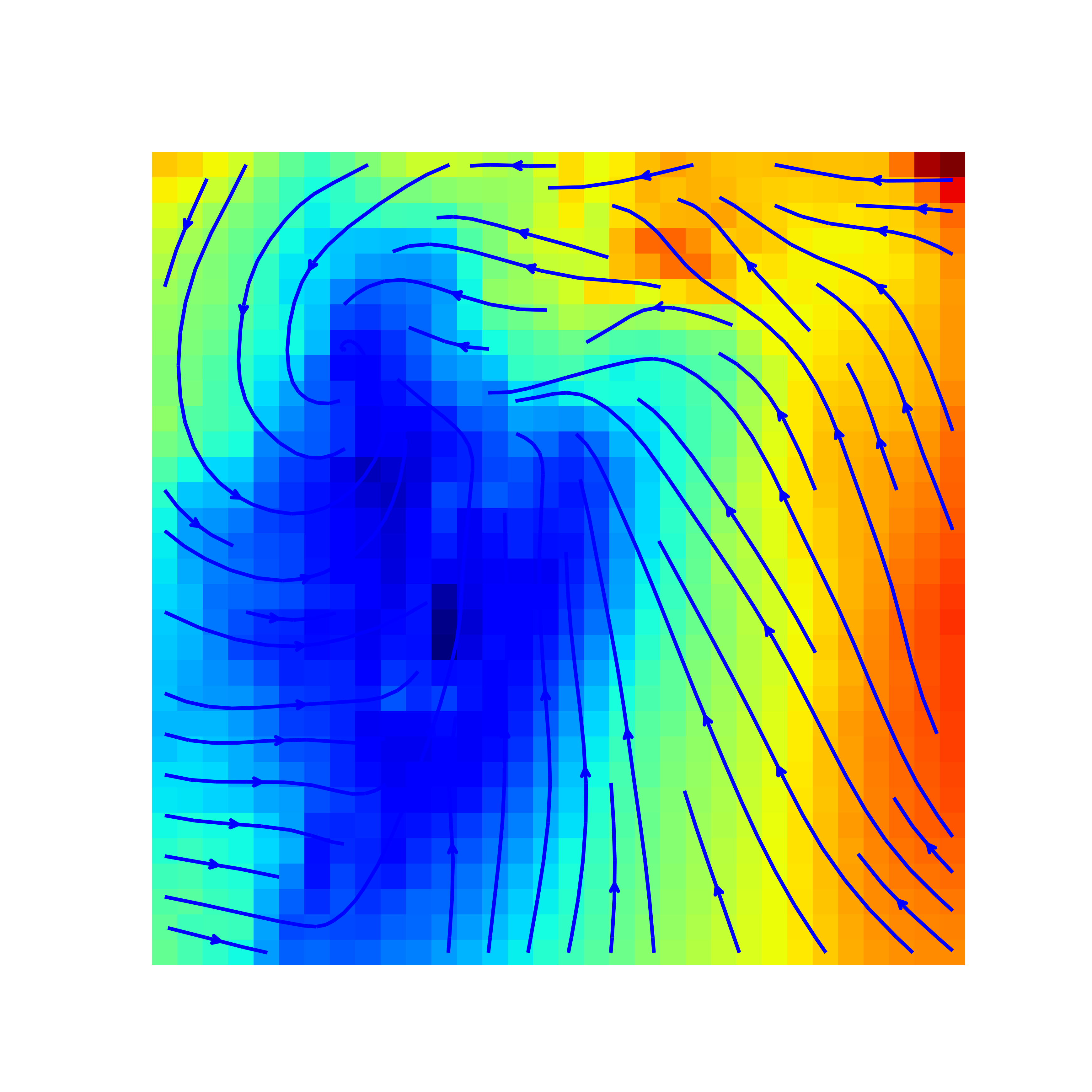}
 \includegraphics[width=0.31\textwidth]{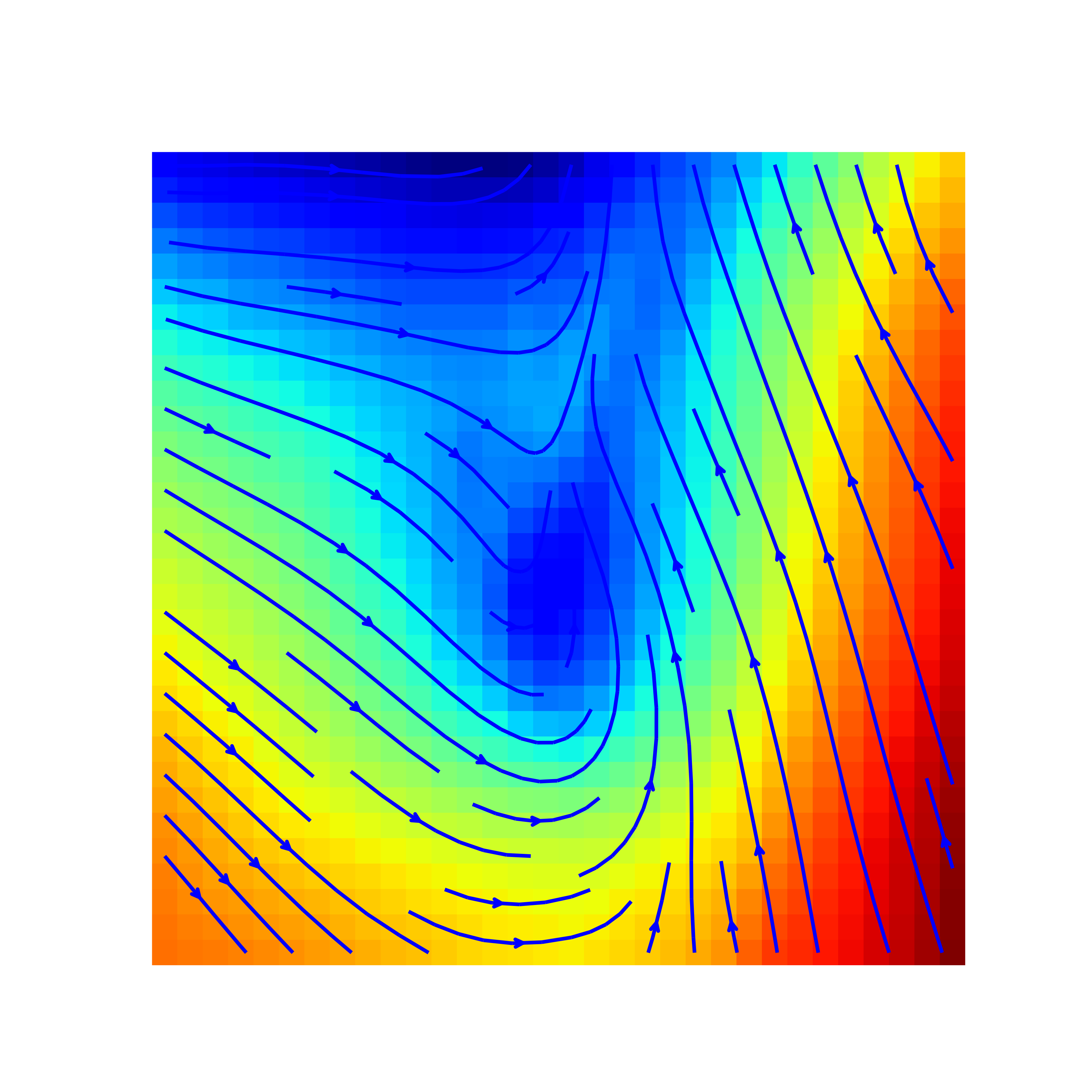}
 \includegraphics[width=0.31\textwidth]{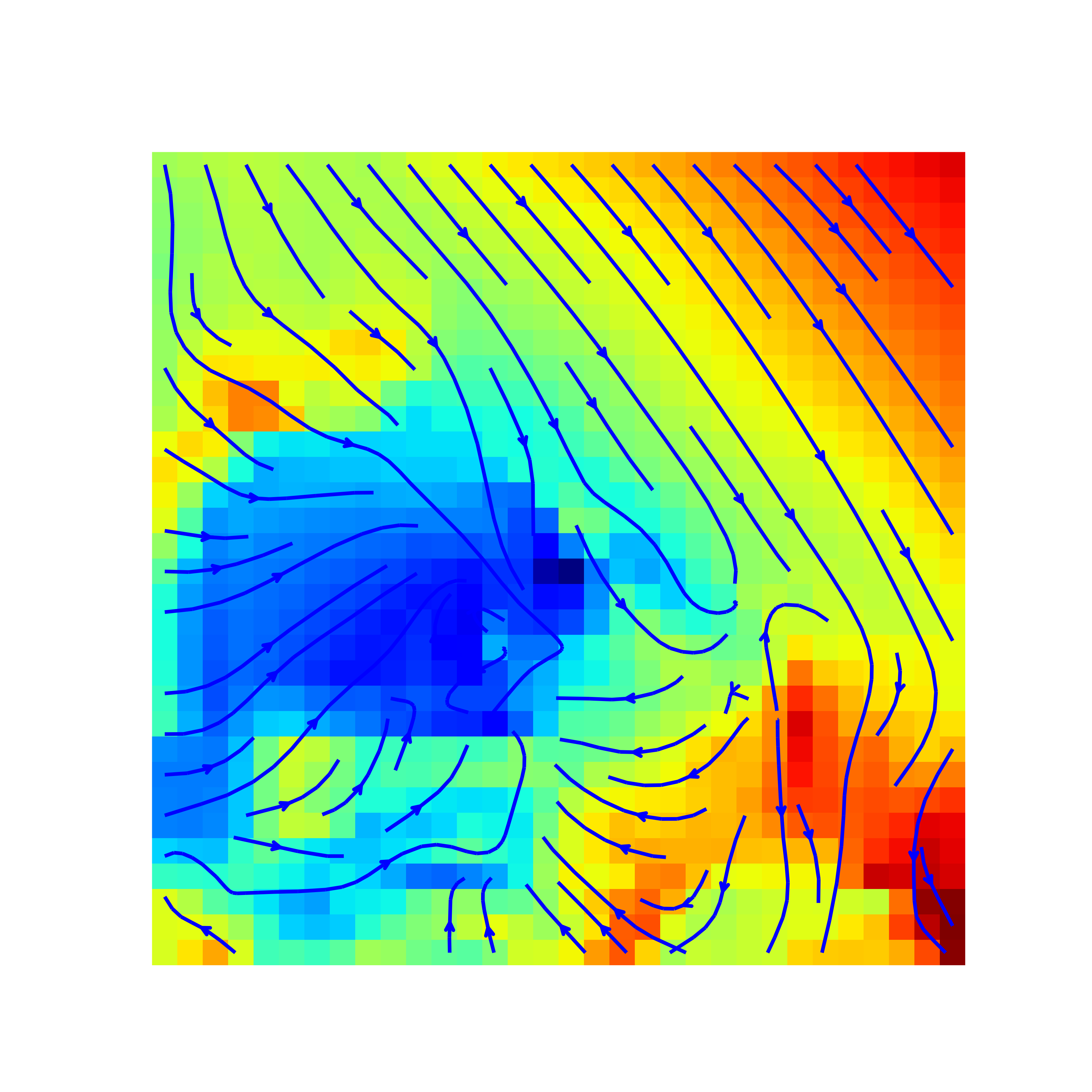}
\end{subfigure}
\caption{Sample images of tropical cyclones mis-classified (false negative) by our deep CNN model. Figure shows sea level pressure (color map) and near surface wind distribution (vector solid line).}
\end{figure}

\subsection{Classification Results for Atmospheric Rivers}
In contrast to tropical cyclones, atmospheric rivers are distinctively different events. They are narrow corridors of concentrated moisture in atmosphere. They usually originate in tropical oceans and move pole-ward. Figure 3 shows examples of correctly classified land falling atmospheric rivers that occur on the western Pacific Ocean and north Atlantic Ocean. The characteristics of narrow water vapor corridor is well defined and clearly observable in these images. 

Figure 4 are examples of mis-classified atmospheric rivers. Upon further investigation, we believe there are two main factors leading to mis-classification. Firstly, presence of weak atmospheric river systems. For instance, the left column of Figure 4 shows comparatively weak atmospheric rivers. The water vapor distribution clearly show a band of concentrated moisture cross mid-latitude ocean, but the signal is much weaker comparing to Figure 3. Thus, deep CNN does not predict them correctly. Secondly, the presence of other climate event may also affect deep CNN representation of atmospheric rivers. In reality, the location and shape of atmospheric river are affected by jet streams and extra-tropical cyclones. For example, Figure 4 right column shows rotating systems (likely extra-tropical cyclone) adjacent to the atmospheric river. This phenomenon presents challenge for deep CNN on representing atmospheric river.

\begin{table}[h]
\noindent
\caption{Confusion matrix for atmospheric river classification}
\begin{tabular}{r|c|c|} 
\multicolumn{1}{r}{} & \multicolumn{1}{c}{Label AR} & \multicolumn{1}{c}{Label Non\_AR} \\
\cline{2-3}
Predict AR & 0.93 & 0.107 \\ 
\cline{2-3}
Predict Non\_AR& 0.07 & 0.893 \\ 
\cline{2-3}
\end{tabular}
\end{table}

\begin{figure}[!h]
\centering
\begin{subfigure}[b]{0.5\textwidth}
  \includegraphics[width=0.5\textwidth]{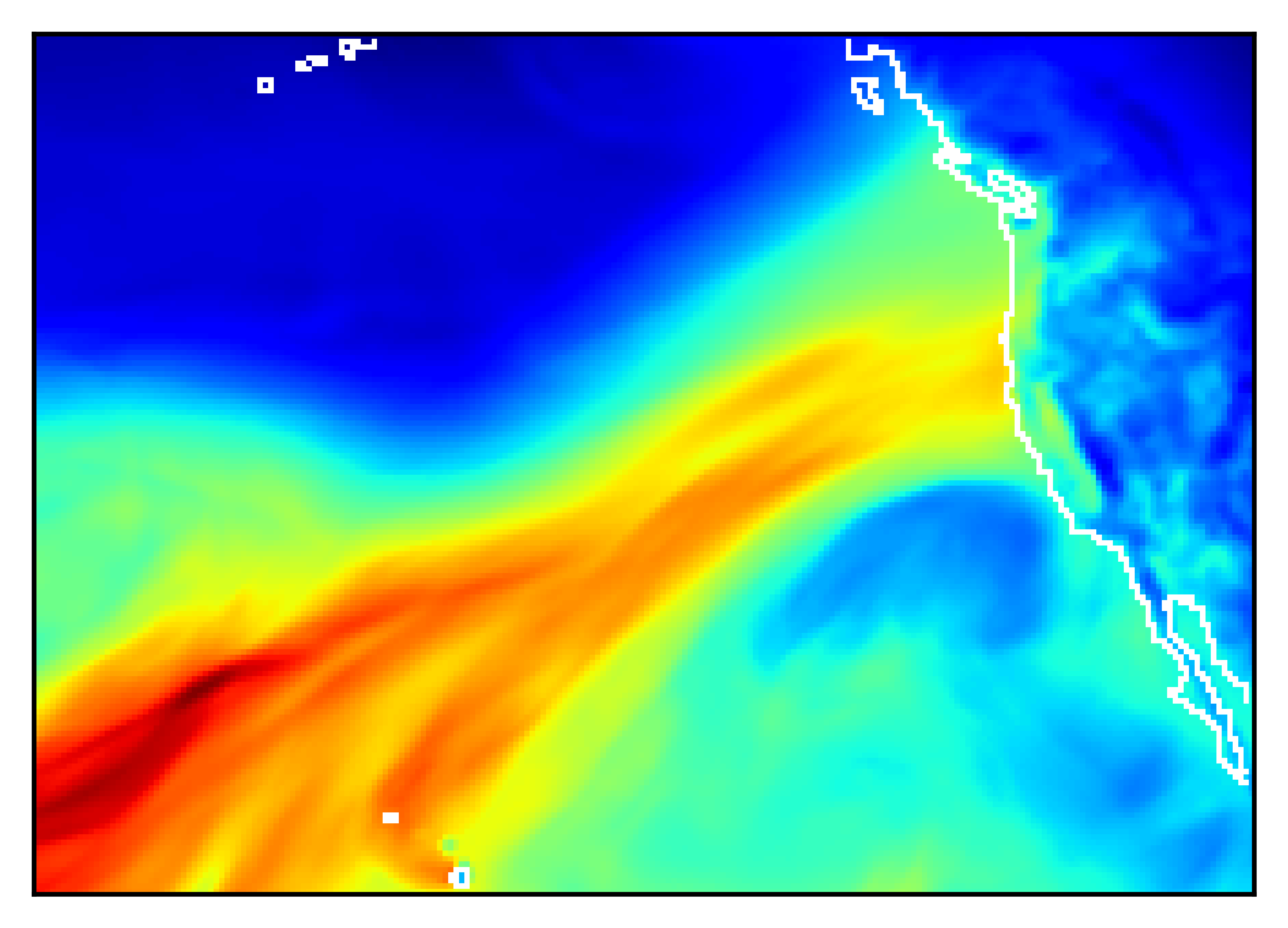}
  \includegraphics[width=0.5\textwidth]{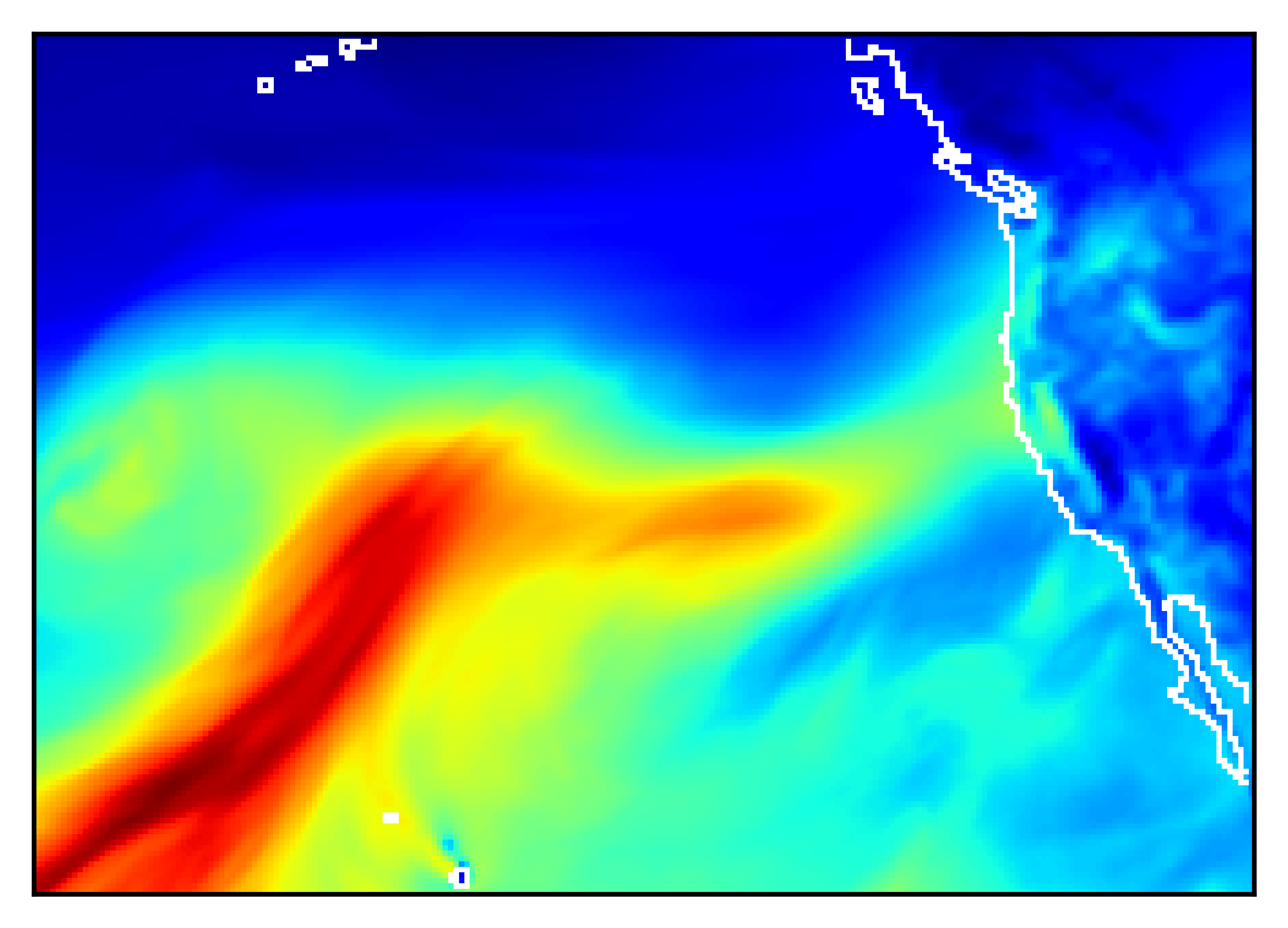}
\end{subfigure}
\begin{subfigure}[b]{0.5\textwidth}
  \includegraphics[width=0.5\textwidth]{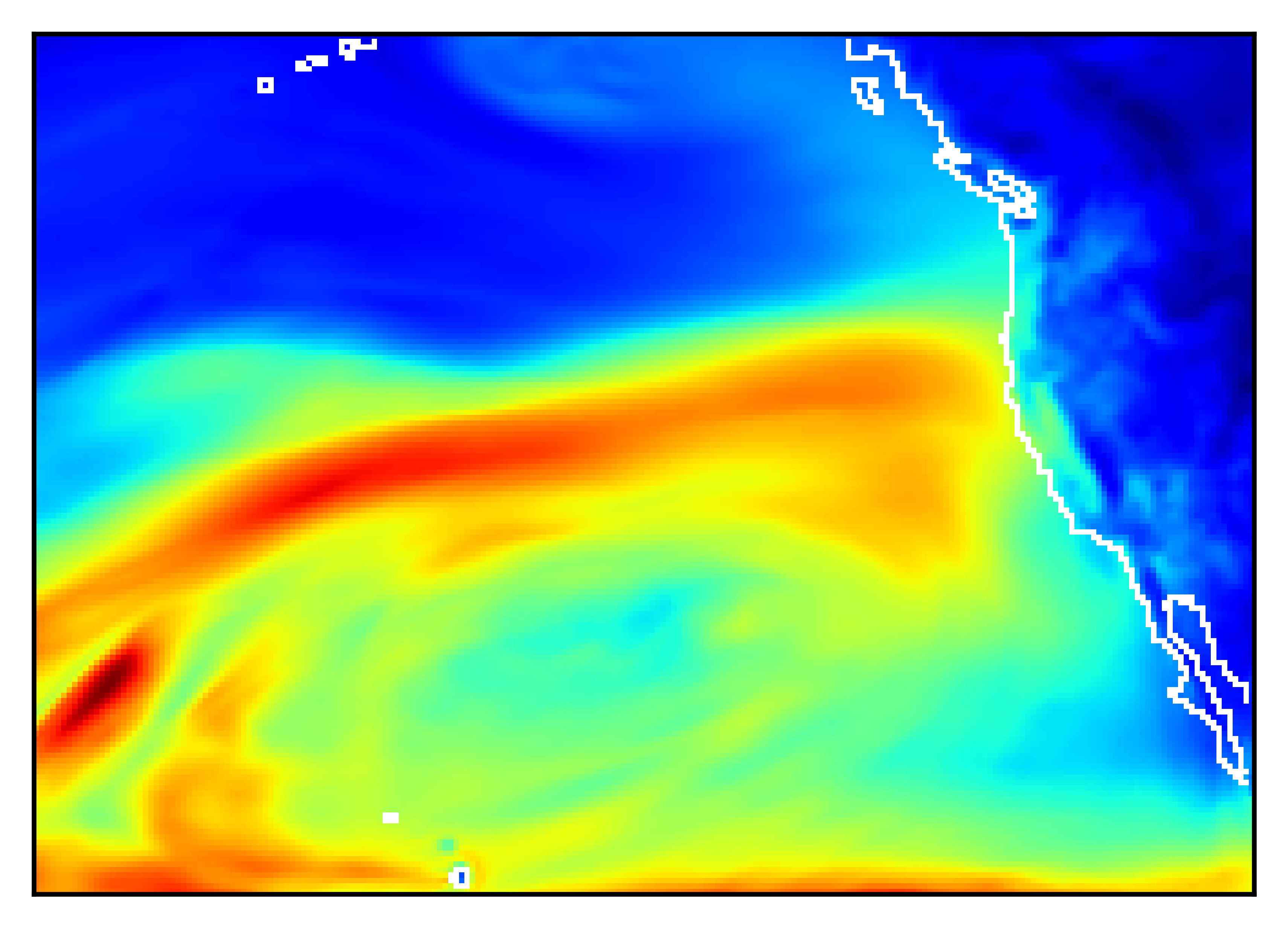}
  \includegraphics[width=0.5\textwidth,height=0.36\textwidth]{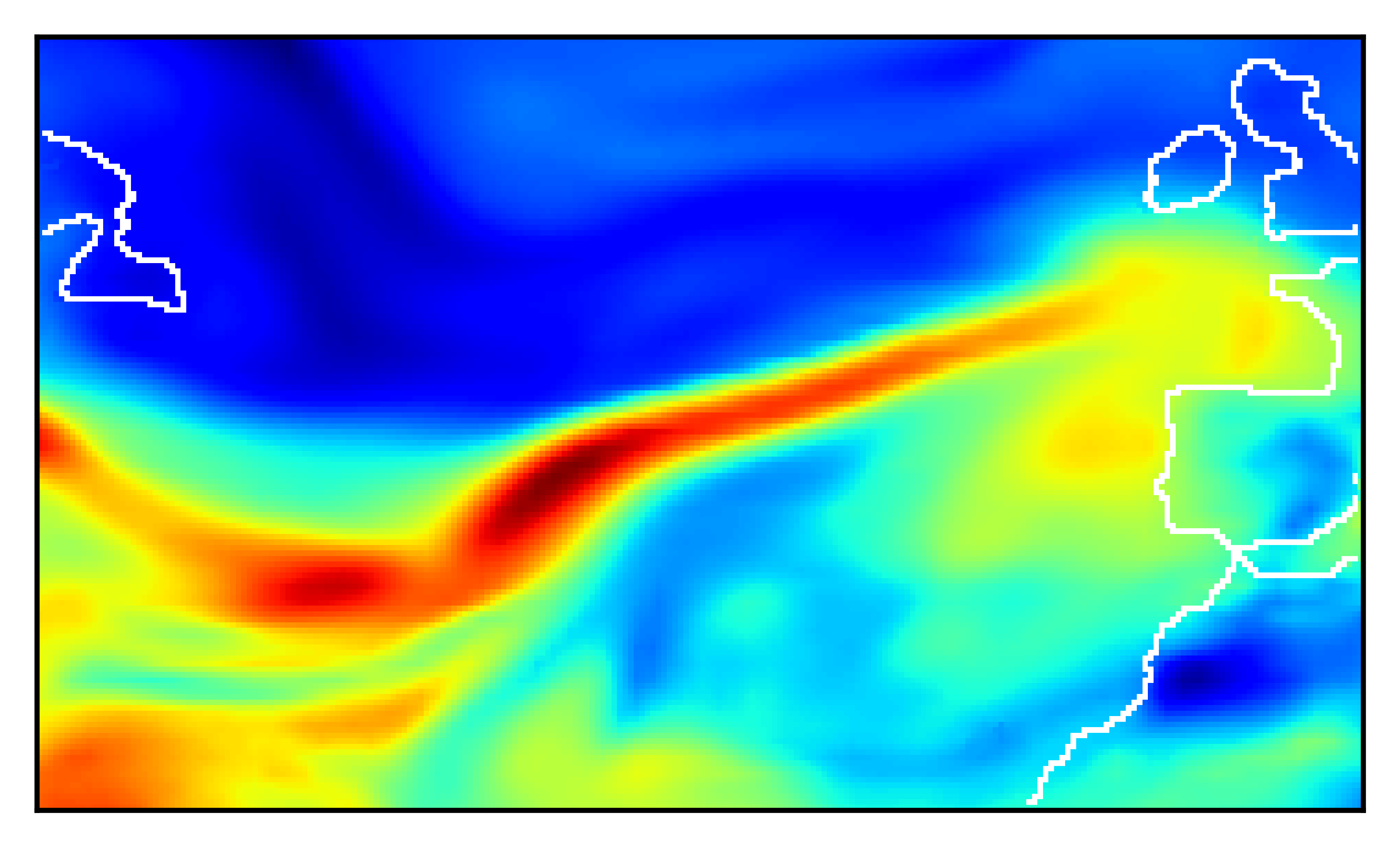}
\end{subfigure} 
\begin{subfigure}[b]{0.5\textwidth}
  \includegraphics[width=0.5\textwidth,height=0.36\textwidth]{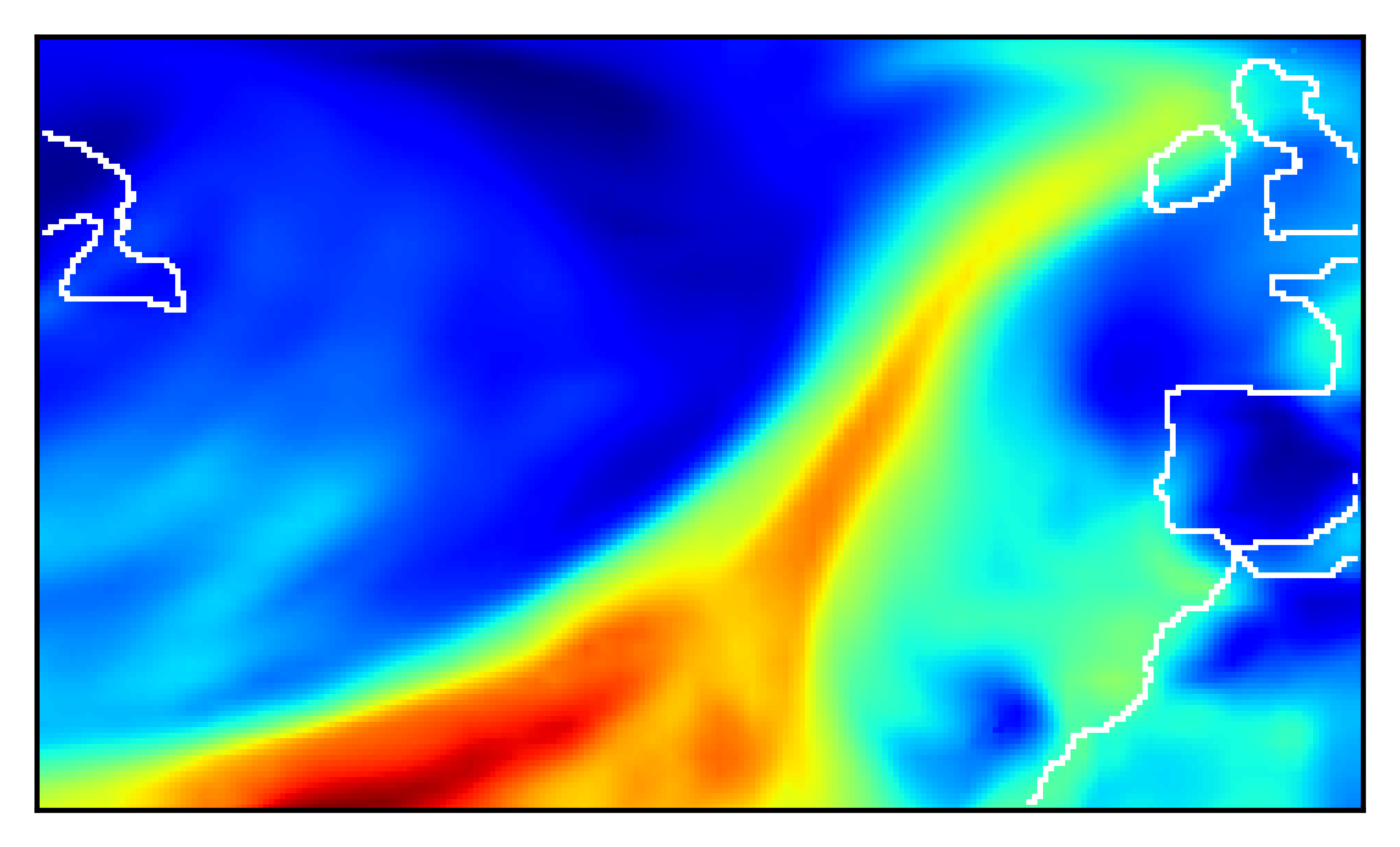}
  \includegraphics[width=0.5\textwidth,height=0.36\textwidth]{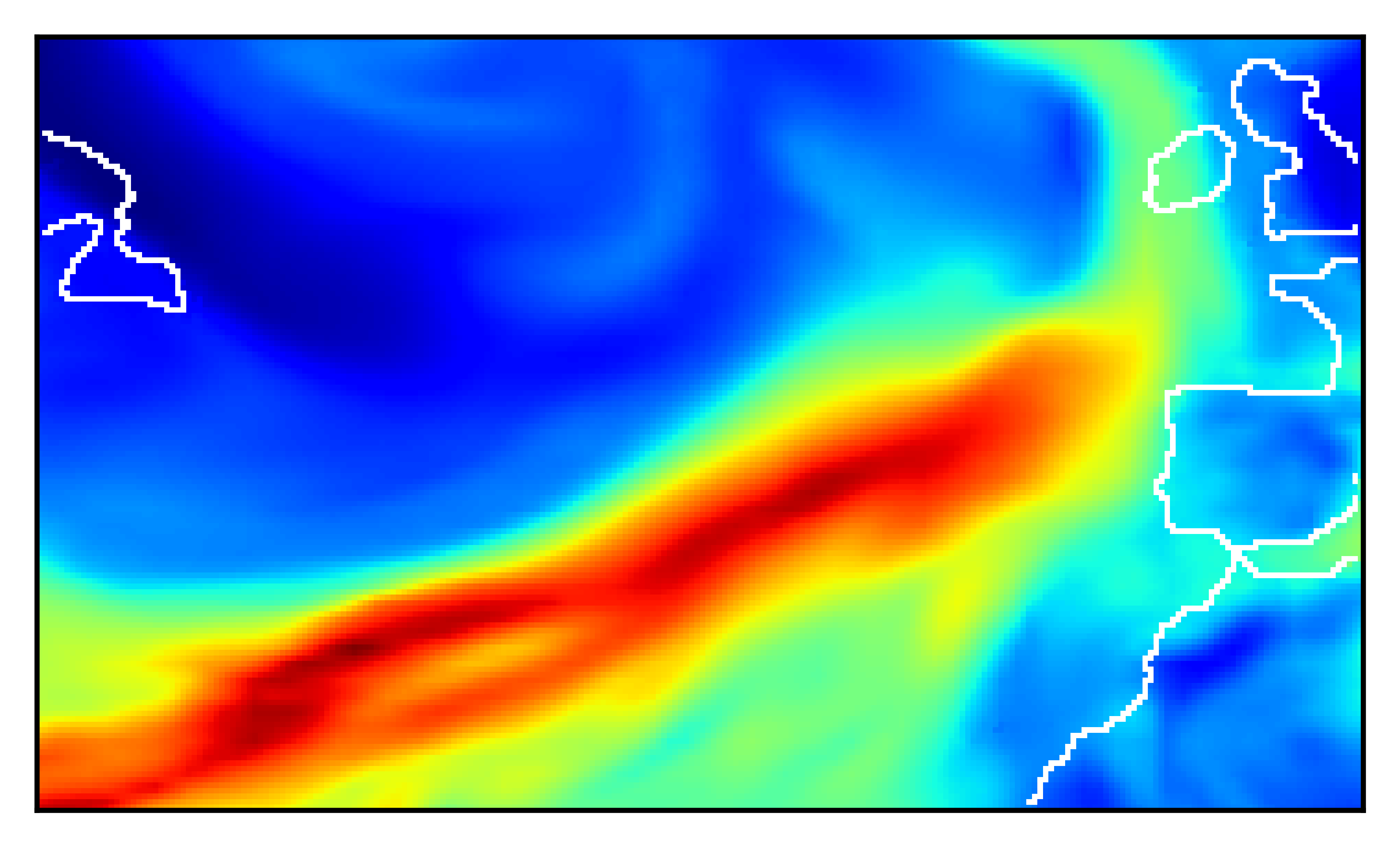}
\end{subfigure} 
\caption{ Sample images of atmospheric rivers correctly classified (true positive) by our deep CNN model. Figure shows total column water vapor (color map) and land sea boundary (solid line).}
\end{figure}

\begin{figure}[!h]
\centering
\begin{subfigure}[b]{0.5\textwidth}
 \includegraphics[width=0.5\textwidth]{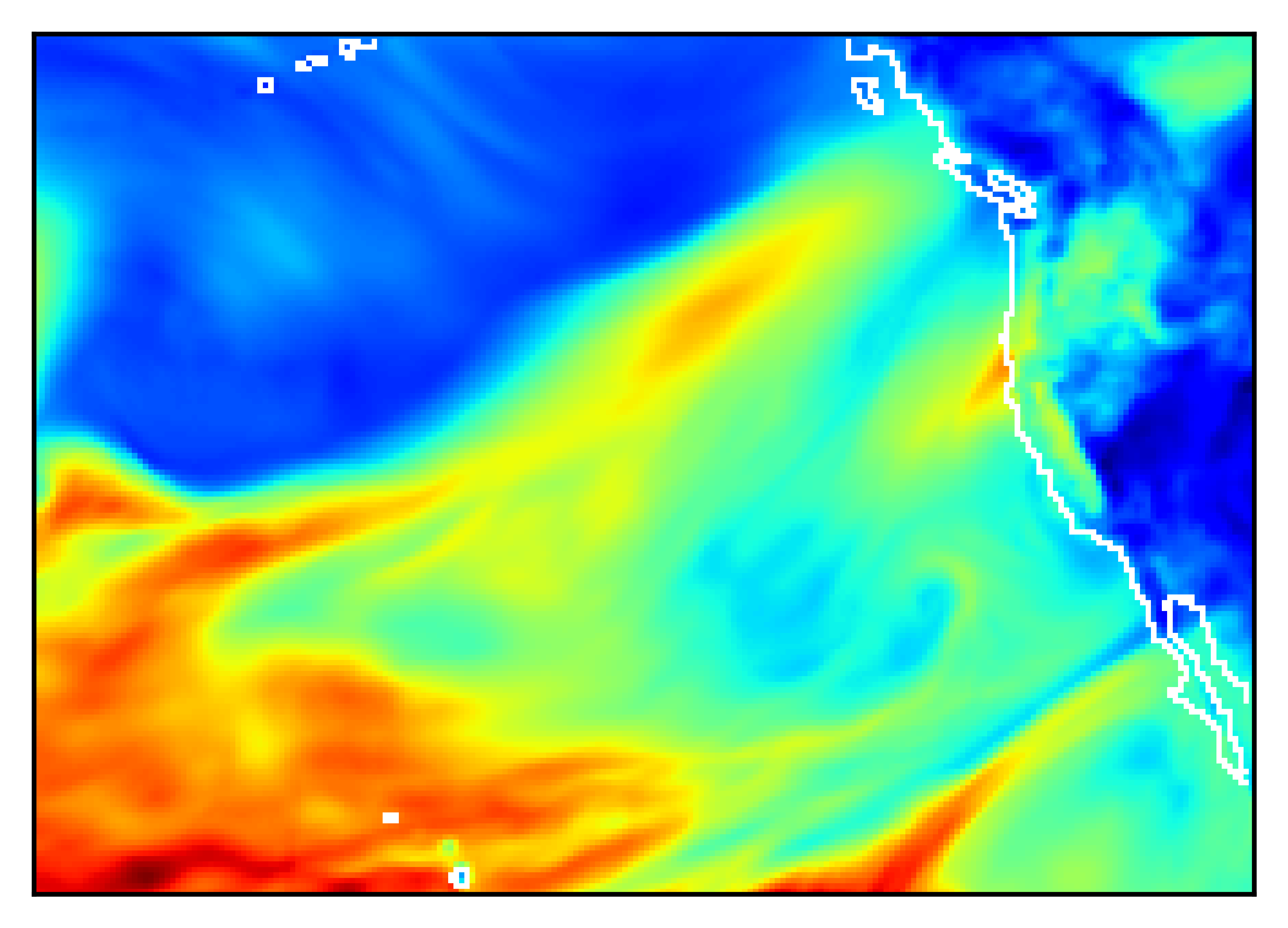}
 \includegraphics[width=0.5\textwidth]{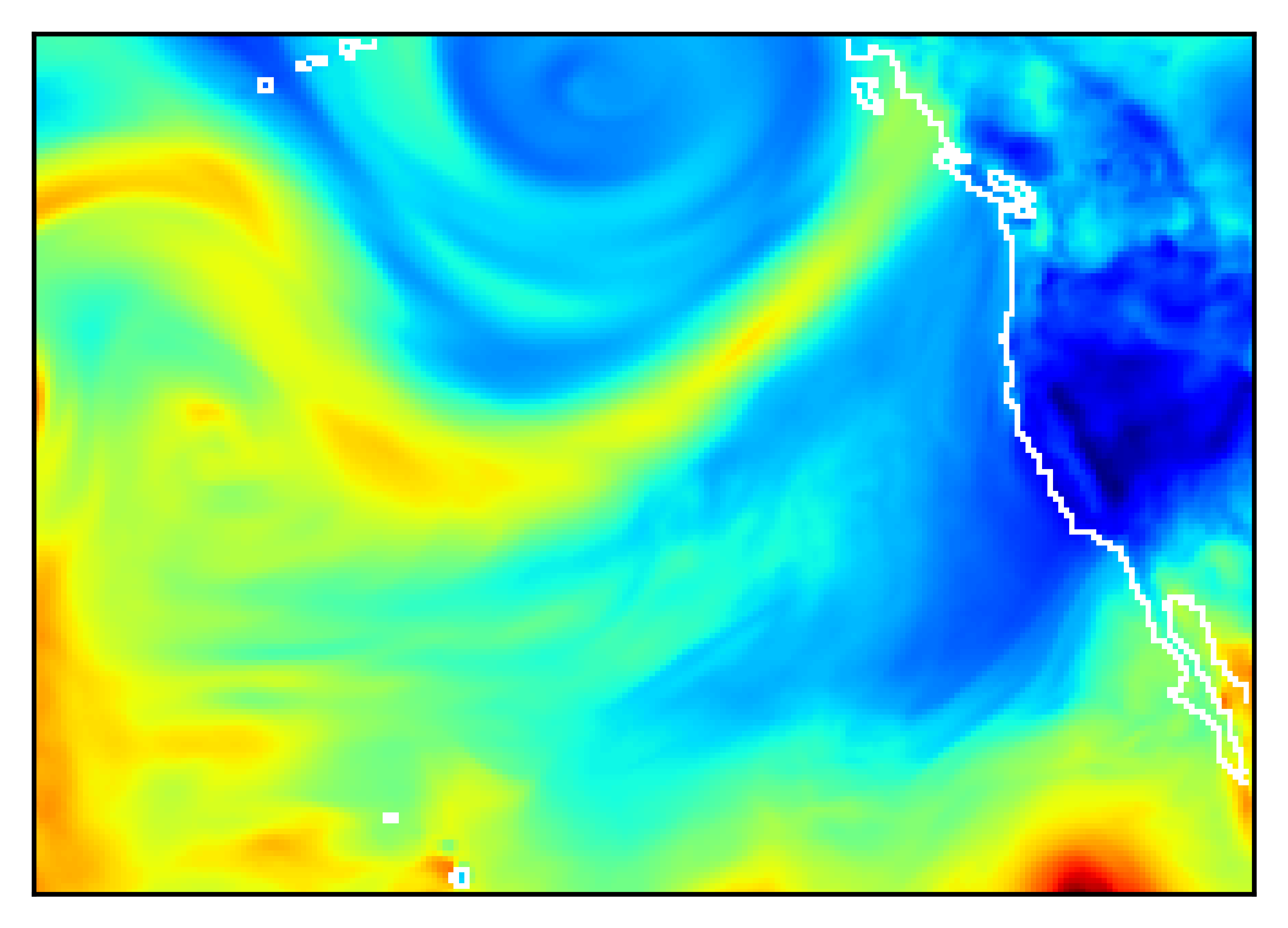}
\end{subfigure}

\begin{subfigure}[b]{0.5\textwidth}
 \includegraphics[width=0.5\textwidth,height=0.36\textwidth]{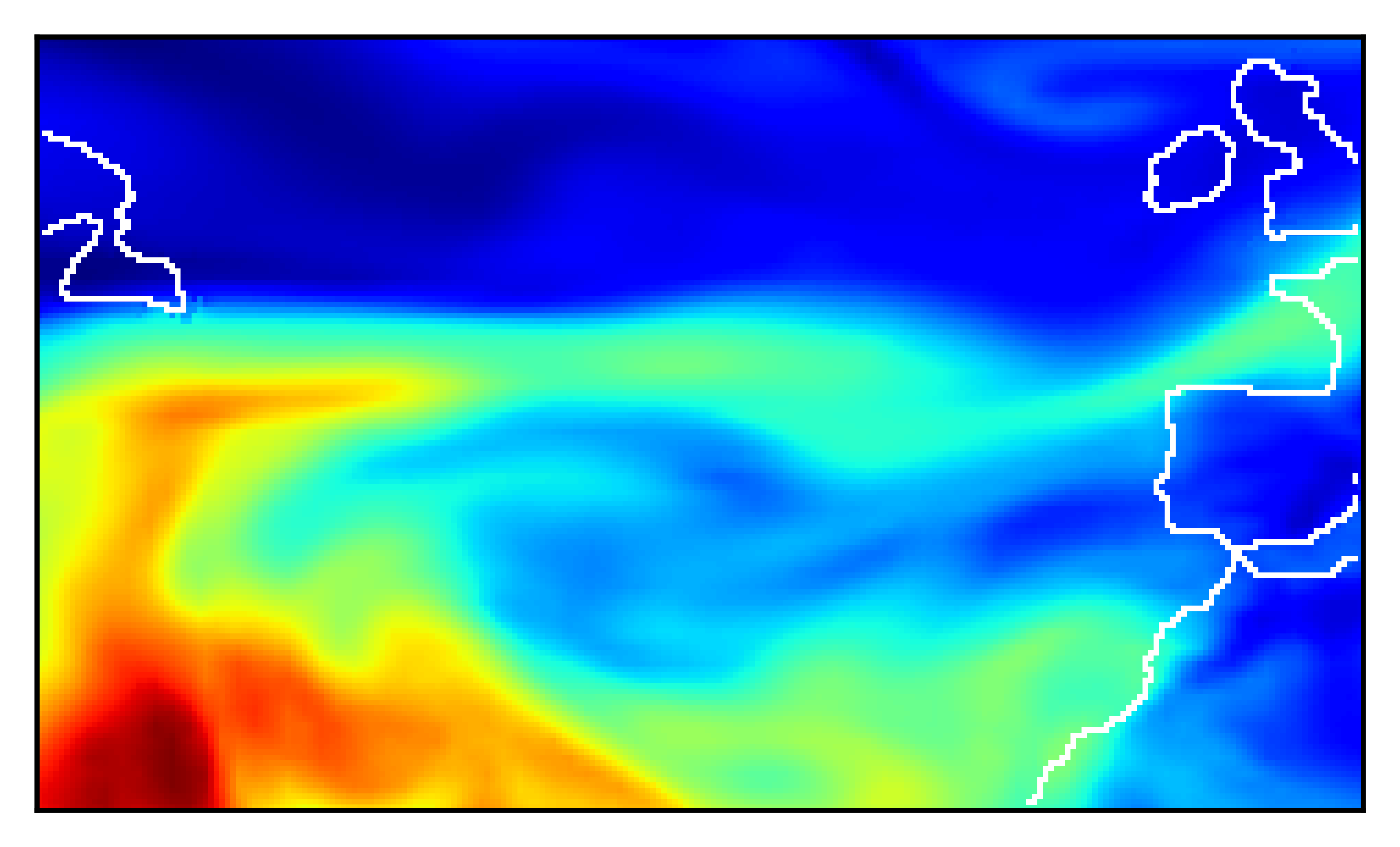}
 \includegraphics[width=0.5\textwidth,height=0.36\textwidth]{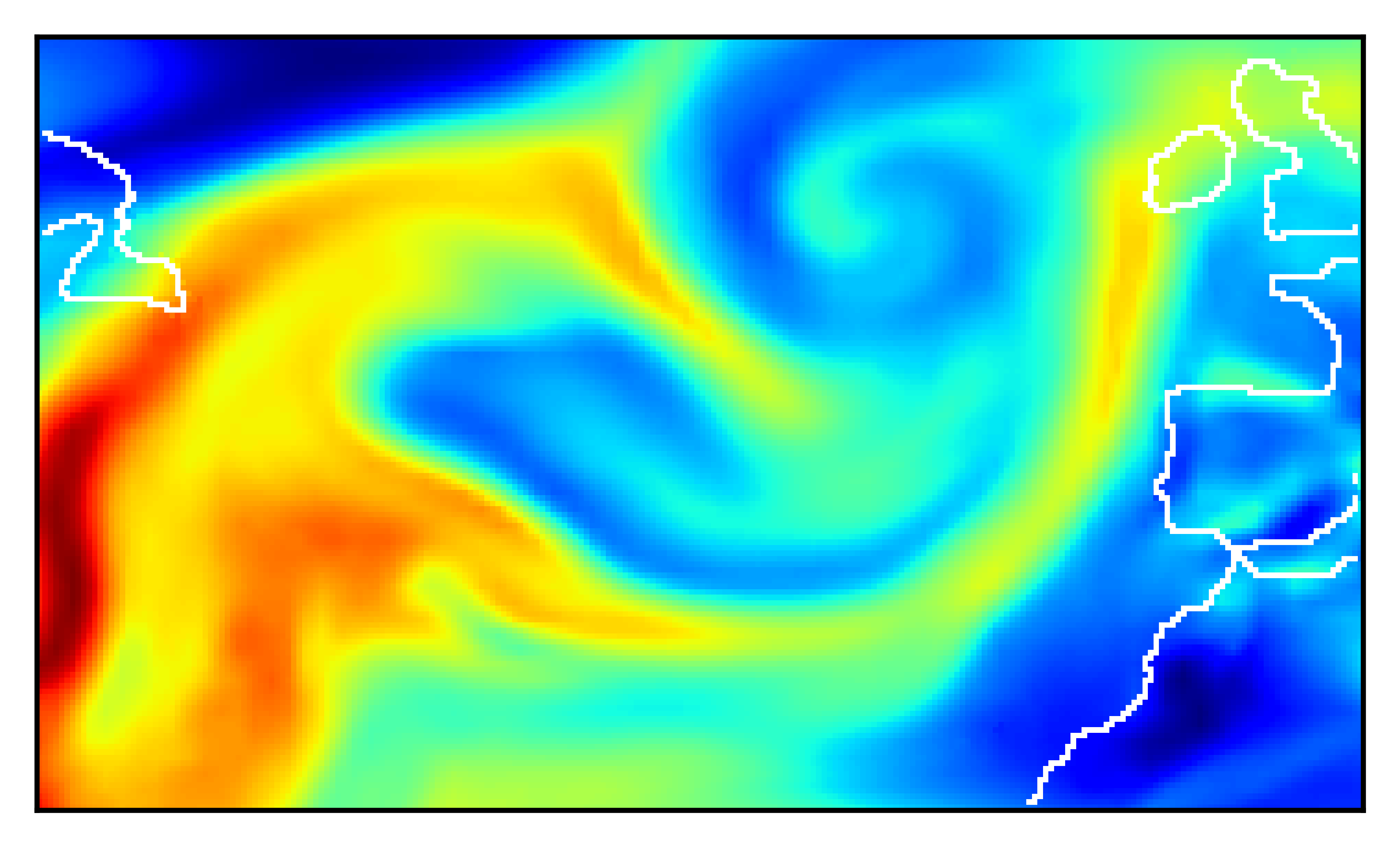}
\end{subfigure}
\caption{Sample images of atmospheric rivers mis-classified (false negative) by our deep CNN model. Figure shows total column water vapor (color map) and land sea boundary (solid line).}
\end{figure}

\subsection{Classification Results for Weather Fronts}
Among the three types of climate events we are looking at, weather fronts have the most complex spatial pattern. Weather fronts typically form at the interface of warm air and cold air, and usually associated with heavy precipitation due moisture condensation of warm air up-lifting. In satellite images,a weather front is observable as a strip of clouds, but it is hardly visible on two dimensional fields such as temperature and pressure. In middle latitude (e.g. most U.S.), a portion of weather front are associated with extra-tropical cyclones. Figure 5 shows examples of correctly classified weather front by our deep CNN system. Visually, the narrow long regions of high precipitation line up approximately parallel to the temperature contour. This is a clear characteristics and comparatively easy for deep CNNs to learn. 

Because patterns of weather fronts is rather complex and hardly show up in two dimensional fields, we decided to further investigate it in later work.

\begin{table}[h]
\noindent
\caption{Confusion matrix for weather front classification}
\begin{tabular}{r|c|c|} 
\multicolumn{1}{r}{} & \multicolumn{1}{c}{Label WF} & \multicolumn{1}{c}{Label Non\_WF} \\
\cline{2-3}
Predict WF & 0.876 & 0.18 \\ 
\cline{2-3}
Predict Non\_WF&0.124 & 0.82 \\ 
\cline{2-3}
\end{tabular}
\end{table}

\begin{figure}[!h]
\centering
\begin{subfigure}[b]{0.5\textwidth}
  \includegraphics[width=0.5\textwidth,height=0.35\textwidth]{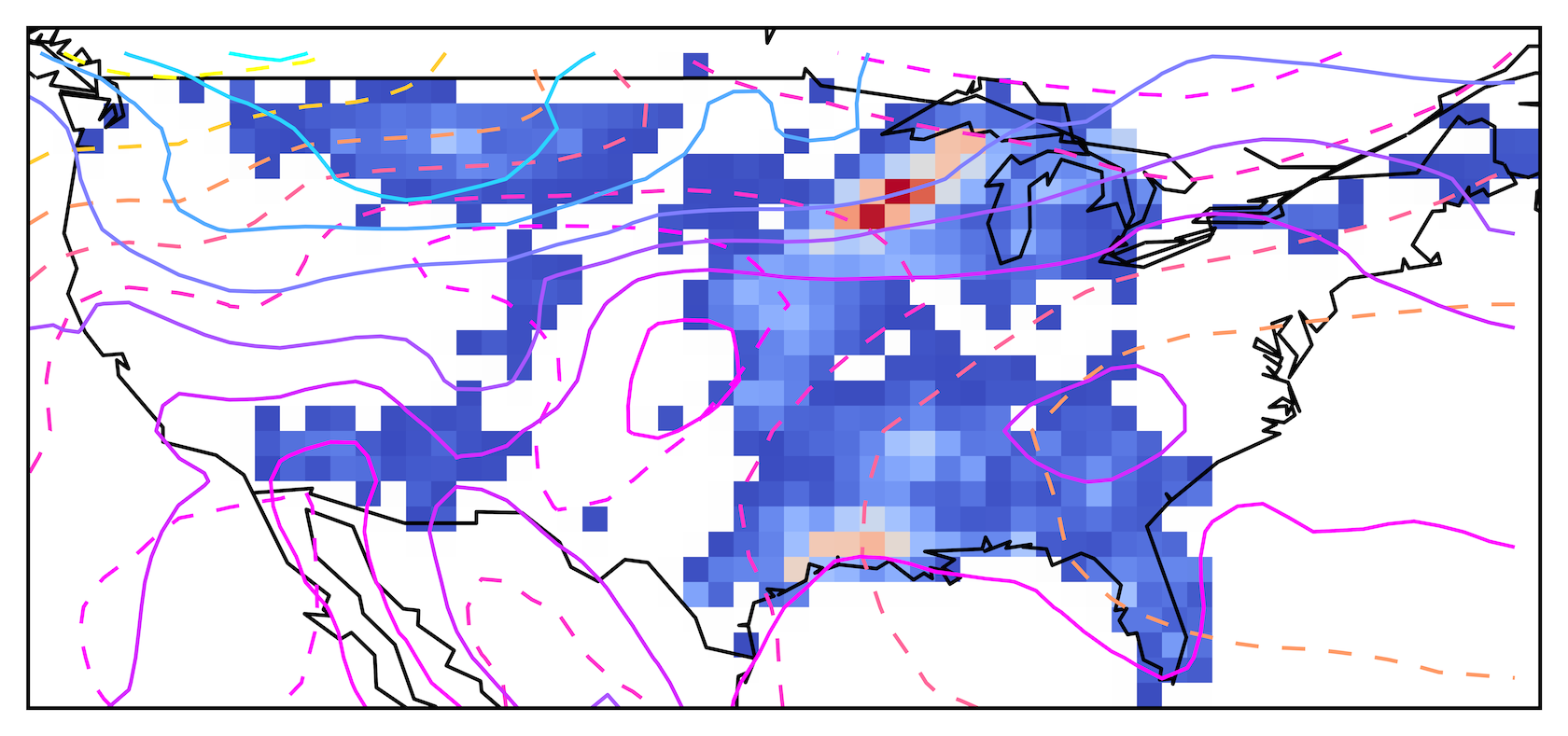}
  \includegraphics[width=0.5\textwidth,height=0.35\textwidth]{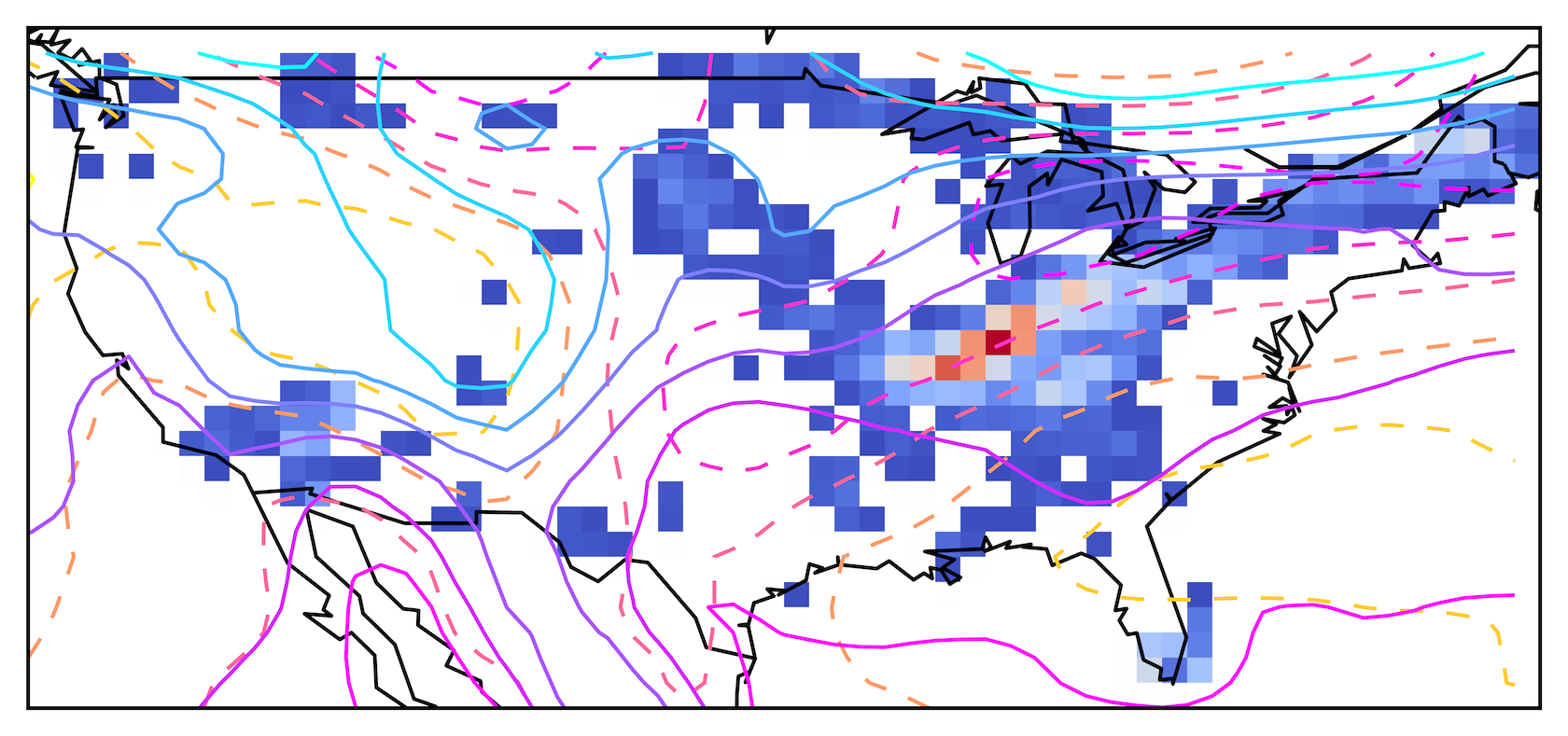}
\end{subfigure}
\begin{subfigure}[b]{0.5\textwidth}
  \includegraphics[width=0.5\textwidth,height=0.35\textwidth]{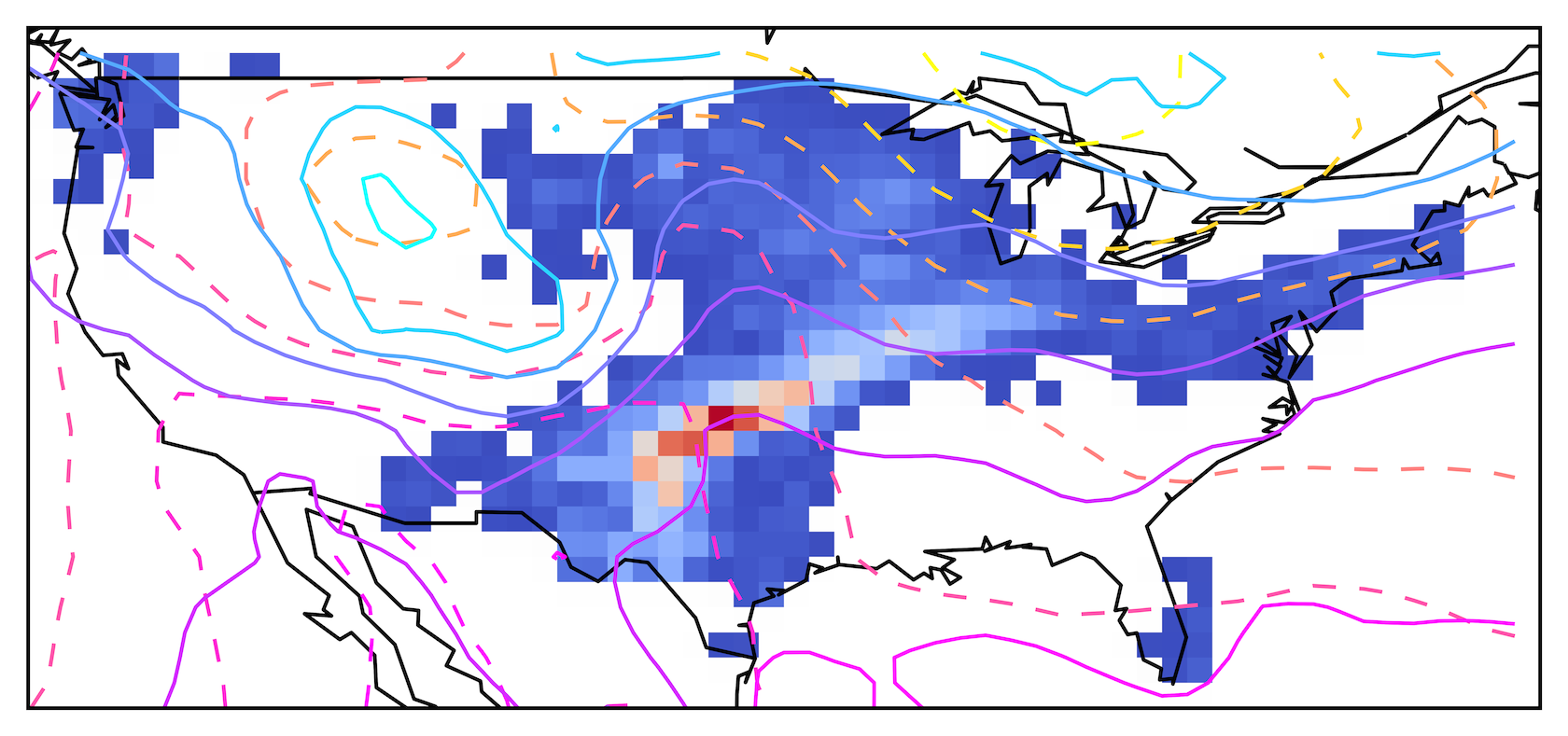}
  \includegraphics[width=0.5\textwidth,height=0.35\textwidth]{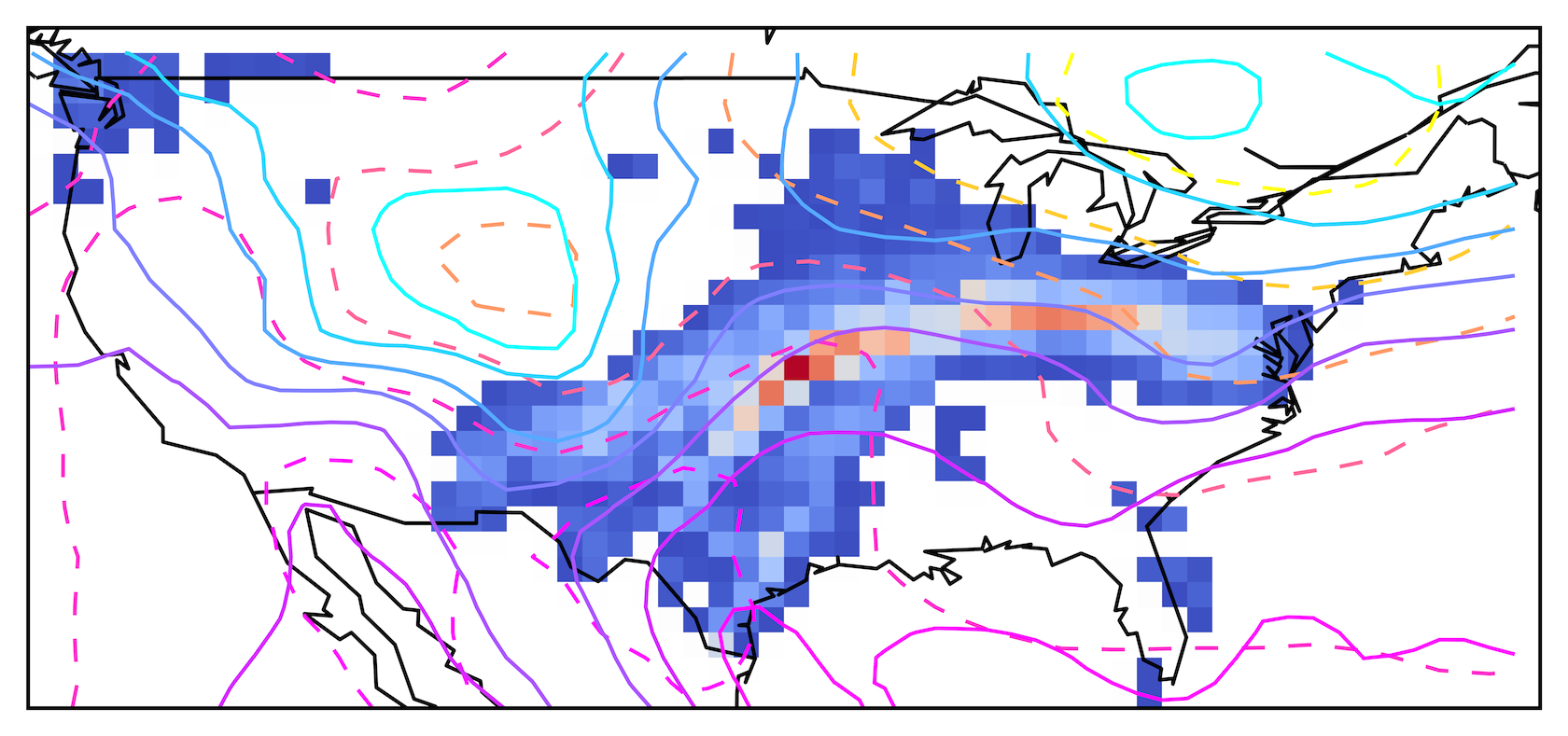}
\end{subfigure} 
\begin{subfigure}[b]{0.5\textwidth}
  \includegraphics[width=0.5\textwidth,height=0.35\textwidth]{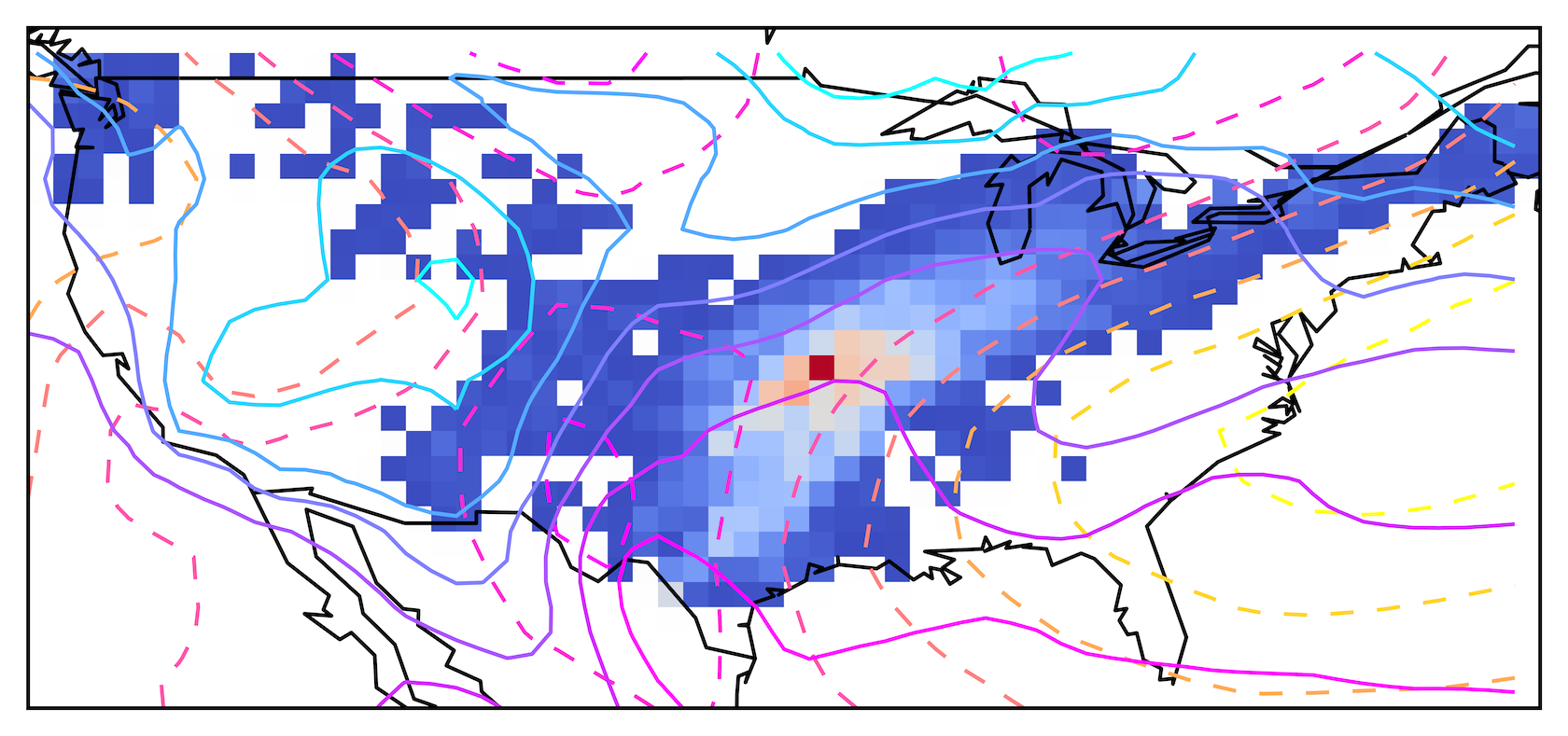}
  \includegraphics[width=0.5\textwidth,height=0.35\textwidth]{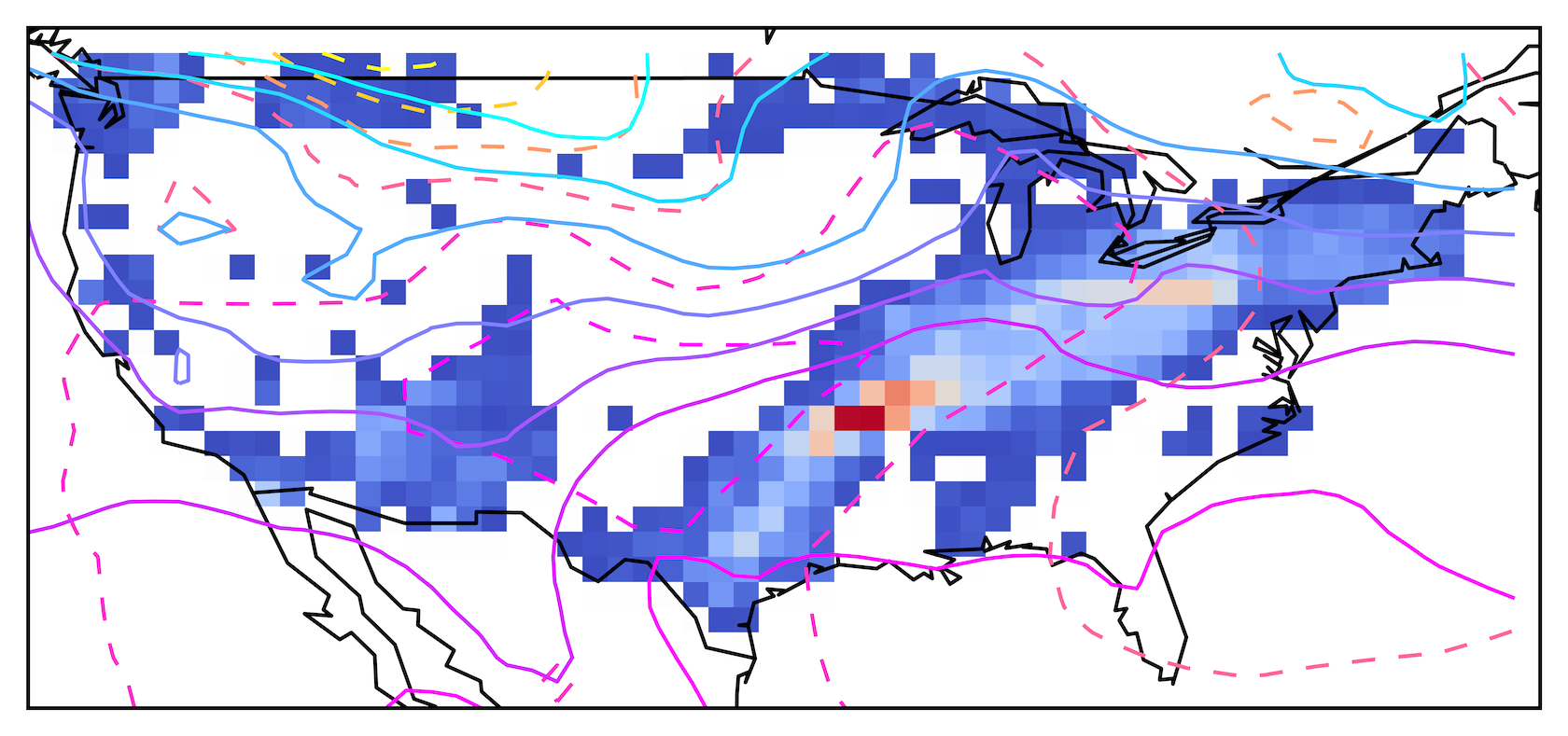}
\end{subfigure} 
\caption{Sample images of weather front correctly classified by our deep CNN model. Figure shows precipitation with daily precipitation less than 5 millimeters filtered out  (color map), near surface air temperature (solid contour line) and sea level pressure (dashed contour line)}
\end{figure}
 
\section{Future Work}
In the present study, we trained deep CNNs separately for classifying tropical cyclones, atmospheric rivers and weather fronts individually. Ideally, we would like to train a \textbf{single} neural network for detecting all three types of events. Unlike object recognition in natural images, climate patterns detection have unique challenges. Firstly, climate events happen at vastly different spatial scales. For example, a tropical cyclone typically extends over less than 500 kilometers in radius, while an atmospheric river can be several thousand kilometers long. Secondly, different climate events are characterized by different sets of physical variables. For example, atmospheric rivers correlate strongly with the vertical integration of water vapor, while tropical cyclones has a more complex multi-variable pattern involving sea level pressure, near surface wind and upper troposphere temperature. Future work will need to develop generative CNN architectures that are capable of discriminating between different variables based on the event type and capable of handling events at various spatial scale. Note that we have primarily addressed \textbf{detection} of extreme weather patterns, but not their \textbf{localization}. We will consider architectures for spatially localizing weather pattern in the future.

Several researchers have pointed out that deeper and larger CNNs perform better for classification and detection tasks\cite{Simonyan14c,szegedy2015going} compared to shallow networks. However, deep networks require huge amount of data to be effectively trained, and to prevent model over fitting. Datasets, such as ImageNet, provide millions of labeled images for training and testing deep and large CNNs. In contrast, we can only obtain a small amount of labeled training data, hence we are constrained on the class of deep CNNs that we can explore without suffering from over-fitting. This limitation also points us to the need for developing unsupervised approaches for climate pattern detection. We believe that this will be critical for the majority of scientific disciplines that typically lack labeled data.

\section{Conclusion}
In this study, we explored deep learning as a methodology for detecting extreme weather patterns in climate data. We developed deep CNN architecture for classifying tropical cyclones, atmospheric rivers and weather fronts. The system achieves fairly high classification accuracy, range from 89\% to 99\%. To the best of our knowledge, this is the first time that deep CNN has been applied to tackle climate pattern recognition problems. This successful application could be a precursor for tackling a broad class of pattern detection problem in climate science. Deep neural network learns high-level representations from data directly, therefore potentially avoiding traditional subjective thresholding based criteria of climate variables for event detection. Results from this study will be used for quantifying climate extreme events trend in current day and future climate scenarios, as well as investigating the changes in dynamics and thermodynamics of extreme events in global warming contend. This information is critical for climate change adaptation, hazard risk prediction and climate change policy making. 

\section{Acknowledgments}
This research was conducted using "Neon", an open source library for deep learning from Nervana Systems. 

This research used resources of the National Energy Research Scientific Computing Center, a DOE Office of Science User Facility supported by the Office of Science of the U.S. Department of Energy under Contract No. DE-AC02-05CH11231. This work was supported by the Director, Office of Science, Office of Advanced Scientific Computing Research, Applied Mathematics program of the U.S. Department of Energy under Contract No. DE-AC02-05CH11231.

\bibliographystyle{abbrv}
\nocite{*}
\bibliography{deeplearning.bib}  

\begin{thebibliography}{10}

\bibitem{brochu2010tutorial}
E.~Brochu, V.~M. Cora, and N.~De~Freitas.
\newblock A tutorial on bayesian optimization of expensive cost functions, with
  application to active user modeling and hierarchical reinforcement learning.
\newblock {\em arXiv preprint arXiv:1012.2599}, 2010.

\bibitem{chattopadhyay2013description}
R.~Chattopadhyay, A.~Vintzileos, and C.~Zhang.
\newblock A description of the madden--julian oscillation based on a
  self-organizing map.
\newblock {\em Journal of Climate}, 26(5):1716--1732, 2013.

\bibitem{dahl2012context}
G.~E. Dahl, D.~Yu, L.~Deng, and A.~Acero.
\newblock Context-dependent pre-trained deep neural networks for
  large-vocabulary speech recognition.
\newblock {\em Audio, Speech, and Language Processing, IEEE Transactions on},
  20(1):30--42, 2012.

\bibitem{girshick2014rich}
R.~Girshick, J.~Donahue, T.~Darrell, and J.~Malik.
\newblock Rich feature hierarchies for accurate object detection and semantic
  segmentation.
\newblock In {\em Proceedings of the IEEE Conference on Computer Vision and
  Pattern Recognition (CVPR)}, pages 580--587, 2014.

\bibitem{glorot2010understanding}
X.~Glorot and Y.~Bengio.
\newblock Understanding the difficulty of training deep feedforward neural
  networks.
\newblock In {\em International conference on artificial intelligence and
  statistics}, pages 249--256, 2010.

\bibitem{gorricha2013framework}
J.~Gorricha, V.~Lobo, and A.~C. Costa.
\newblock A framework for exploratory analysis of extreme weather events using
  geostatistical procedures and 3d self-organizing maps.
\newblock {\em International Journal on Advances in Intelligent Systems}, 6(1),
  2013.

\bibitem{graves2013speech}
A.~Graves, A.-r. Mohamed, and G.~Hinton.
\newblock Speech recognition with deep recurrent neural networks.
\newblock In {\em Acoustics, Speech and Signal Processing (ICASSP), 2013 IEEE
  International Conference on}, pages 6645--6649. IEEE, 2013.

\bibitem{hinton2012deep}
G.~Hinton, L.~Deng, D.~Yu, G.~E. Dahl, A.-r. Mohamed, N.~Jaitly, A.~Senior,
  V.~Vanhoucke, P.~Nguyen, T.~N. Sainath, et~al.
\newblock Deep neural networks for acoustic modeling in speech recognition: The
  shared views of four research groups.
\newblock {\em Signal Processing Magazine, IEEE}, 29(6):82--97, 2012.

\bibitem{iglesiasexamination}
G.~Iglesias, D.~C. Kale, and Y.~Liu.
\newblock An examination of deep learning for extreme climate pattern analysis.
\newblock In {\em The 5th International Workshop on Climate Informatics}, 2015.

\bibitem{krizhevsky2012imagenet}
A.~Krizhevsky, I.~Sutskever, and G.~E. Hinton.
\newblock Imagenet classification with deep convolutional neural networks.
\newblock In {\em Advances in Neural Information Processing Systems (NIPS)},
  pages 1097--1105, 2012.

\bibitem{kunkel2012meteorological}
K.~E. Kunkel, D.~R. Easterling, D.~A. Kristovich, B.~Gleason, L.~Stoecker, and
  R.~Smith.
\newblock Meteorological causes of the secular variations in observed extreme
  precipitation events for the conterminous united states.
\newblock {\em Journal of Hydrometeorology}, 13(3):1131--1141, 2012.

\bibitem{larochelle2009exploring}
H.~Larochelle, Y.~Bengio, J.~Louradour, and P.~Lamblin.
\newblock Exploring strategies for training deep neural networks.
\newblock {\em The Journal of Machine Learning Research}, 10:1--40, 2009.

\bibitem{lavers2012detection}
D.~A. Lavers, G.~Villarini, R.~P. Allan, E.~F. Wood, and A.~J. Wade.
\newblock The detection of atmospheric rivers in atmospheric reanalyses and
  their links to british winter floods and the large-scale climatic
  circulation.
\newblock {\em Journal of Geophysical Research: Atmospheres}, 117(D20), 2012.

\bibitem{lecun1998gradient}
Y.~LeCun, L.~Bottou, Y.~Bengio, and P.~Haffner.
\newblock Gradient-based learning applied to document recognition.
\newblock {\em Proceedings of the IEEE}, 86(11):2278--2324, 1998.

\bibitem{nair2010rectified}
V.~Nair and G.~E. Hinton.
\newblock Rectified linear units improve restricted boltzmann machines.
\newblock In {\em Proceedings of the 27th International Conference on Machine
  Learning (ICML)}, pages 807--814, 2010.

\bibitem{Nolan2012tropical}
D.~S. Nolan and M.~G. McGauley.
\newblock Tropical cyclogenesis in wind shear: Climatological relationships and
  physical processes.
\newblock In {\em Cyclones: Formation, Triggers, and Control}, pages 1--36.
  Nova Science Publishers, 2012.

\bibitem{Prabhat2015teca}
Prabhat, S.~Byna, V.~Vishwanath, E.~Dart, M.~Wehner, W.~D. Collins, et~al.
\newblock Teca: Petascale pattern recognition for climate science.
\newblock In {\em Computer Analysis of Images and Patterns}, pages 426--436.
  Springer, 2015.

\bibitem{Prabhat2012teca}
Prabhat, O.~R{\"u}bel, S.~Byna, K.~Wu, F.~Li, M.~Wehner, W.~Bethel, et~al.
\newblock Teca: A parallel toolkit for extreme climate analysis.
\newblock In {\em Third Worskhop on Data Mining in Earth System Science (DMESS)
  at the International Conference on Computational Science (ICCS)}, 2012.

\bibitem{ruhmelhart1986learning}
D.~Ruhmelhart, G.~Hinton, and R.~Wiliams.
\newblock Learning representations by back-propagation errors.
\newblock {\em Nature}, 323:533--536, 1986.

\bibitem{sermanet14over}
P.~Sermanet, D.~Eigen, X.~Zhang, M.~Mathieu, R.~Fergus, and Y.~LeCun.
\newblock Overfeat: Integrated recognition, localization and detection using
  convolutional networks.
\newblock In {\em International Conference on Learning Representations (ICLR)},
  2014.

\bibitem{shi2015convolutional}
X.~Shi, Z.~Chen, H.~Wang, D.-Y. Yeung, W.-K. Wong, and W.-c. Woo.
\newblock Convolutional lstm network: A machine learning approach for
  precipitation nowcasting.
\newblock In {\em Advances in Neural Information Processing Systems:
  Twenty-Ninth Annual Conference on Neural Information Processing Systems
  (NIPS)}, 2015.

\bibitem{Simonyan14c}
K.~Simonyan and A.~Zisserman.
\newblock Very deep convolutional networks for large-scale image recognition.
\newblock In {\em Internaltional Conference on Learning Representation (ICLR)},
  2015.

\bibitem{SnoekGit}
J.~Snoek.
\newblock Spearmint.
\newblock \url{https://github.com/HIPS/Spearmint}, 2015.

\bibitem{snoek2012practical}
J.~Snoek, H.~Larochelle, and R.~P. Adams.
\newblock Practical bayesian optimization of machine learning algorithms.
\newblock In {\em Advances in neural information processing systems}, pages
  2951--2959, 2012.

\bibitem{snoek2015scalable}
J.~Snoek, O.~Rippel, K.~Swersky, R.~Kiros, N.~Satish, N.~Sundaram, M.~Patwary,
  M.~Prabhat, and R.~Adams.
\newblock Scalable bayesian optimization using deep neural networks.
\newblock In {\em Proceedings of The 32nd International Conference on Machine
  Learning}, pages 2171--2180, 2015.

\bibitem{SprDosBroRied15}
J.~T. Springenberg, A.~Dosovitskiy, T.~Brox, and M.~Riedmiller.
\newblock Striving for simplicity: The all convolutional net.
\newblock In {\em International Conference on Learning Representation (ICLR)},
  2015.

\bibitem{sutskever2014sequence}
I.~Sutskever, O.~Vinyals, and Q.~V. Le.
\newblock Sequence to sequence learning with neural networks.
\newblock In {\em Advances in neural information processing systems}, pages
  3104--3112, 2014.

\bibitem{szegedy2015going}
C.~Szegedy, W.~Liu, Y.~Jia, P.~Sermanet, S.~Reed, D.~Anguelov, D.~Erhan,
  V.~Vanhoucke, and A.~Rabinovich.
\newblock Going deeper with convolutions.
\newblock In {\em Proceedings of the IEEE Conference on Computer Vision and
  Pattern Recognition (CVPR)}, pages 1--9, 2015.

\bibitem{uijlings2013selective}
J.~R. Uijlings, K.~E. van~de Sande, T.~Gevers, and A.~W. Smeulders.
\newblock Selective search for object recognition.
\newblock {\em International Journal of Computer Vision}, 104(2):154--171,
  2013.

\bibitem{vitart1997simulation}
F.~Vitart, J.~Anderson, and W.~Stern.
\newblock Simulation of interannual variability of tropical storm frequency in
  an ensemble of gcm integrations.
\newblock {\em Journal of Climate}, 10(4):745--760, 1997.

\bibitem{vitart1999impact}
F.~Vitart, J.~Anderson, and W.~Stern.
\newblock Impact of large-scale circulation on tropical storm frequency,
  intensity, and location, simulated by an ensemble of gcm integrations.
\newblock {\em Journal of Climate}, 12(11):3237--3254, 1999.

\bibitem{walsh2007objectively}
K.~Walsh, M.~Fiorino, C.~Landsea, and K.~McInnes.
\newblock Objectively determined resolution-dependent threshold criteria for
  the detection of tropical cyclones in climate models and reanalyses.
\newblock {\em Journal of Climate}, 20(10):2307--2314, 2007.

\bibitem{walsh1997tropical}
K.~Walsh and I.~G. Watterson.
\newblock Tropical cyclone-like vortices in a limited area model: comparison
  with observed climatology.
\newblock {\em Journal of Climate}, 10(9):2240--2259, 1997.

\bibitem{wehner2015resolution}
M.~Wehner, Prabhat, K.~A. Reed, D.~Stone, W.~D. Collins, and J.~Bacmeister.
\newblock Resolution dependence of future tropical cyclone projections of cam5.
  1 in the us clivar hurricane working group idealized configurations.
\newblock {\em Journal of Climate}, 28(10):3905--3925, 2015.

\end{thebibliography}
\end{document}